\documentclass[10pt,twocolumn,letterpaper]{article}

\usepackage[sort, numbers]{natbib}
\usepackage{tikz,pgfplots}
\usepackage{titling}
\pgfplotsset{compat=newest}
\usepackage{iccv}
\usepackage{times}
\usepackage{epsfig}
\usepackage{graphicx}
\usepackage{amsmath}
\usepackage{amssymb}
\usepackage{epsfig}
\usepackage{caption}
\usepackage{bigdelim}
\usepackage{dsfont}
\usepackage{makecell}
\usepackage{vcell}
\usepackage{enumitem}
\setitemize{noitemsep,topsep=0pt,parsep=0pt,partopsep=0pt}
\usepackage{rotating}
\usepackage{mathtools}
\usepackage{filecontents}
\usetikzlibrary{decorations.pathreplacing,calligraphy}
\usepackage{float}
\usepackage{pdfpages}
\usepackage[toc,page]{appendix}
\usepackage{multirow}
\usepackage{amsthm}
\usepackage[ruled,vlined]{algorithm2e}
\usepackage[noend]{algpseudocode}
\usepackage[list=true]{subcaption}
\usepgfplotslibrary{groupplots}
\newcommand{\centered}[1]{\begin{tabular}{l} #1 \end{tabular}}
\usepackage{xcolor}

\DeclarePairedDelimiterX{\norm}[1]{\lVert}{\rVert}{#1}

\newcommand{\loss}{\mathcal{L}(\mathbb{P}_{\mathbf{x}|\mathbf{y}},\Pggiveny)}

\newcommand{\expect}[2]{\mathbb{E}_{#1}\left[#2\right]}

\newcommand{\Pxgiveny}{\mathbb{P}_{\mathbf{x}|\mathbf{y}}}
\newcommand{\Pxgivenyeq}{\mathbb{P}_{\mathbf{x}|\mathbf{y}=y}}

\newcommand{\Px}{\mathbb{P}_{\mathbf{x}}}

\newcommand{\Pz}{\mathbb{P}_{\mathbf{z}}}

\newcommand{\Pggiveny}[1][\boldy]{\mathbb{P}_{\mathbf{g}_{\theta}|#1}}
\newcommand{\Pggivenyeq}{\mathbb{P}_{\mathbf{g}_{\theta}|\boldy=y}}

\newcommand{\boldx}{\mathbf{x}}

\newcommand{\boldy}{\mathbf{y}}
\newcommand{\boldz}{\mathbf{z}}
\newcommand{\boldn}{\mathbf{n}}
\newcommand{\gtheta}{\mathbf{g}_{\theta}}

\newcommand{\supf}[1][\boldy]{\sup_{f_{#1}\in L_{1}}}

\newcommand{\real}[4]{\centered{\includegraphics[width=0.105\textwidth]{images/ours/#1-#2/#3/#4/real.png}}}
\newcommand{\stddev}[5]{\centered{\includegraphics[width=0.105\textwidth]{images/ours/#1-#2/#3/#4/std_dev_z#5.png}}}
\newcommand{\noisy}[4]{\centered{\includegraphics[width=0.105\textwidth]{images/ours/#1-#2/#3/#4/noisy.png}}}
\newcommand{\oursa}[4]{\centered{\includegraphics[width=0.105\textwidth]{images/ours/#1-#2/#3/#4/mean.png}}}

\newcommand{\ourss}[5]{\centered{\includegraphics[width=0.105\textwidth]{images/ours/#1-#2/#3/#4/#5fake_z1.0.png}}}
\newcommand{\lag}[5]{\centered{\includegraphics[width=0.105\textwidth]{images/ours-lag/#1-#2/#3/#4/#5fake_z1.0.png}}}
\newcommand{\laglikely}[4]{\centered{\includegraphics[width=0.105\textwidth]{images/ours-lag/#1-#2/#3/#4/00fake_z0.0.png}}}
\newcommand{\dncnn}[2]{\centered{\includegraphics[width=0.105\textwidth]{images/dncnn/#1/#2/denoised.png}}}
\newcommand{\oursmse}[2]{\centered{\includegraphics[width=0.105\textwidth]{images/ours-mse/#1/#2/denoised.png}}}
\newcommand{\oursslikely}[5]{\centered{\includegraphics[width=0.105\textwidth]{images/ours/#1-#2/#3/#4/#5fake_z0.75.png}}}

\makeatletter
\newcommand{\subalign}[1]{%
  \vcenter{%
    \Let@ \restore@math@cr \default@tag
    \baselineskip\fontdimen10 \scriptfont\tw@
    \advance\baselineskip\fontdimen12 \scriptfont\tw@
    \lineskip\thr@@\fontdimen8 \scriptfont\thr@@
    \lineskiplimit\lineskip
    \ialign{\hfil$\m@th\scriptstyle##$&$\m@th\scriptstyle{}##$\hfil\crcr
      #1\crcr
    }%
  }%
}
\makeatother

\usepackage{array, blindtext}

\usepackage[english]{babel}
\usepackage[pagebackref=true,breaklinks=true,letterpaper=true,colorlinks,bookmarks=false]{hyperref}
\addto\extrasenglish{
    
}
\addto\extrasenglish{
    
}

\iccvfinalcopy 

\def\iccvPaperID{3} 

\begin{document}
\title{High Perceptual Quality Image Denoising with a Posterior Sampling CGAN}

\author{
    Guy Ohayon \\ Technion
    \and
    Theo Adrai \\ Technion
    \and
    Gregory Vaksman \\ Technion
    \and
    Michael Elad \\ Google Research
    \and
    Peyman Milanfar \\ Google Research
}
\maketitle
\begin{abstract}
The vast work in Deep Learning (DL) has led to a leap in image denoising research.
Most DL solutions for this task have chosen to put their efforts on the denoiser's architecture while maximizing distortion performance.
However, distortion driven solutions lead to blurry results with sub-optimal perceptual quality, especially in immoderate noise levels.
In this paper we propose a different perspective, aiming to produce sharp and visually pleasing denoised images that are still faithful to their clean sources.
Formally, our goal is to achieve high perceptual quality with acceptable distortion.
This is attained by a stochastic denoiser that samples from the posterior distribution, trained as a generator in the framework of conditional generative adversarial networks (CGAN).
Contrary to distortion-based regularization terms that conflict with perceptual quality, we introduce to the CGAN objective a theoretically founded penalty term that does not force a distortion requirement on individual samples, but rather on their mean.
We showcase our proposed method with a novel denoiser architecture that achieves the reformed denoising goal and produces vivid and diverse outcomes in immoderate noise levels.
\end{abstract}
\section{Introduction}
Image denoising is one of the most fundamental problems in image processing, and as such it has been explored quite extensively.
As deep learning emerged in the past decade, many neural-network-based attempts were made to solve this task.
These led to state-of-the-art (SoTA) performance in commonly used full reference \emph{distortion} measures, such as Mean-Squared-Error (MSE), that quantify the discrepancy between the denoised image and its clean source~\cite{Zhang_DnCnn_2017, Zhang_FFDNet_2018, Vaksman_2020_CVPR_Workshops, denoising_comaprison, TIAN2020251, yu2019deep}.
While optimizing a distortion measure leads to denoised images that are faithful to their clean sources, the \emph{perceptual quality}, which is the degree to which a denoised image looks natural, is also an important measure to consider.
Recent works have tried to achieve higher perceptual quality compared to distortion based solutions by concurrently optimizing both measures~\cite{Dey2020,Divakar_2017_CVPR_Workshops}.
However, these attempts achieve sub-optimal perceptual quality~\cite{Blau_2018}, and to the best of our knowledge, there are hardly any deep learning solutions that address the image denoising problem while targeting optimal perceptual performance.

In this work we seek to obtain a denoiser that achieves very high perceptual quality, while accompanied by a guarantee on its distortion performance.
As shown in~\cite{Blau_2018}, sampling from the posterior distribution achieves such a goal, compromising 3dB on the optimal Peak Signal To Noise Ratio (PSNR) performance, making such a stochastic denoiser an excellent candidate for our needs.
The idea of using posterior sampling for solving image restoration tasks has already been suggested in various contexts~\cite{song2021scorebased, Tonolini2020VariationalIF,adler2018deep}.
Although high dimensional posterior sampling is still considered as a challenging task, recent deep learning methods seem to provide practical tools for handling it.

The success of generative adversarial networks (GAN) has led authors to incorporate sampling (not necessarily from the posterior distribution) to solve various image restoration tasks on certain classes of images~\cite{Bahat_2020_CVPR, Menon_2020_CVPR, berthelot2020creating, whang2020approximate}, and to excellent sampling capabilities from class-specific priors~\cite{Karras_2020_CVPR, Karras_2019_CVPR}.
Most of these were possible due to the improvements in the generative adversarial learning scheme~\cite{gans_comparison, wgan, gulrajani2017improved} that allowed stable training, contrary to the instabilities of the originally proposed GAN optimization objective~\cite{gan}.
The authors of~\cite{adler2018deep} have shown that the CGAN objective~\cite{mirza2014conditional} formalized under the Wasserstein-1 metric~\cite{gulrajani2017improved,wgan} theoretically drives a conditional generator to sample from the posterior distribution.
Therefore, such an optimization framework provides a practical way to approximate the desired sampling.
For instance, the Latent Adversarial Generator (LAG)~\cite{berthelot2020creating} has shown SoTA single image super resolution (SISR) results from an extremely low resolution input, attained by a tweaked version of CGAN.

Rather than seeking a balance between perceptual quality and distortion performance, we aim to sample from the posterior distribution while willing to compromise up to 3dB in PSNR performance.
In order to regularize the proposed sampling, we leverage the property that any stochastic denoiser that samples from such a distribution must also agree in expectation with it.
We introduce a term to the CGAN objective that penalizes solutions which do not satisfy such a necessary property.
Unlike other related methods, our regularization term does not force a distortion requirement on individual denoised samples, but rather on their mean.
Our proposed denoiser's architecture is a novel encoder-decoder, inspired by StyleGAN2~\cite{Karras_2020_CVPR} and UNet~\cite{ronneberger2015unet}, with a high receptive field and a noise injection scheme generalizing that of StyleGAN~\cite{Karras_2019_CVPR}.
We showcase the capabilities of our proposed method in high noise conditions, basing our experiments on several data sets.
\section{Proposed Method: Derivations}\label{theoretical}
Assume an unknown distribution of images $\Px$ and a known stochastic degradation operator $\mathbf{deg(\cdot)}$ (such as additive Gaussian noise). Our goal is to sample from the posterior distribution $\Pxgiveny$ with the help of an independent random vector $\boldz$ of known distribution.
We assume that given ${\boldy=\mathbf{deg(\boldx)}}$, a degraded observation of $\boldx$, there exists a parametric mapping $\gtheta=G_{\theta}(\boldz,\boldy)$ such that ${\boldz\sim\Pz,\gtheta|\boldy\sim\Pxgiveny}$, and $\Pz$ is a known latent distribution where $\boldz$ and $\boldy$ are mutually independent.

For a given $\boldy=y$ (denoting $y$ as a realization of the random variable $\boldy$), the Wasserstein-1 distance~\cite{wgan} between $\Pxgivenyeq$ and $\Pggivenyeq$ can be shown to satisfy the equality
\begin{equation}\label{loss:w1}
\begin{gathered}
    W_{1}(\Pxgivenyeq,\Pggivenyeq)\\=\supf[]\expect{\boldx|\boldy}{f(\boldx,y)}-\expect{\gtheta|\boldy}{f(\gtheta,y)},
\end{gathered}
\end{equation}
where $L_{1}$ is the set of all functions ${f:\mathcal{X}\times\mathcal{Y}\rightarrow\mathbb{R}}$ that are 1-Lipschitz on $\mathcal{X}$ for any $y\in\mathcal{Y}$.
Observe that (\ref{loss:w1}) is defined for a given realization of $\boldy$, whereas we seek an optimization objective that considers all possible ones.
This can be accomplished by taking an expectation on both sides with respect to $\boldy$. The work in~\cite{adler2018deep} shows that such an expectation taken on the right hand side in (\ref{loss:w1}) commutes with the supremum, leading to
\begin{equation}\label{loss:revised_w1}
\begin{gathered}
    \expect{\boldy}{W_{1}(\Pxgiveny,\Pggiveny)}\\=\supf[]\expect{\boldx,\boldy}{f(\boldx,\boldy)}-\expect{\gtheta,\boldy}{f(\gtheta,\boldy)}.
\end{gathered}
\end{equation}
Observe that contrary to (\ref{loss:w1}), the expectations in (\ref{loss:revised_w1}) act on the joint distributions $\mathbb{P}_{\boldx,\boldy}$ and $\mathbb{P}_{\gtheta,\boldy}$.
Therefore, assuming that the function $f$ in the supremum of (\ref{loss:revised_w1}) can be found 
for each $y$, one could evaluate this distance 
as follows:
\begin{itemize}[leftmargin=*]
    \item Draw samples of $\boldx\sim\Px$ (e.g., get an image data set).
    \item Perform $\boldy=\mathbf{deg}(\boldx)$ on each sample of $\boldx$, to obtain samples of $\boldy$ (e.g., contaminate 
    with noise).
    \item Draw independently samples of $\boldz\sim\Pz$.
    \item Compute $G_{\theta}(\boldz,\boldy)$ on each sample of $\boldy$ and $\boldz$, to obtain samples of $\gtheta$ (e.g., denoise each noisy image with a stochastic denoiser).
    \item We now have samples drawn from both $\mathbb{P}_{\boldx,\boldy}$ and $\mathbb{P}_{\gtheta,\boldy}$. Evaluate (\ref{loss:revised_w1}) using the law of large numbers.
\end{itemize}
Considering $G_{\theta}(\boldz,\boldy)$ as a generator and assuming that $f$ is somehow 
realized for each $\theta$, we could optimize for $\theta$:
\begin{equation}\label{obj:cgan}
\begin{gathered}
\min_{\theta}\loss\\
    =\min_{\theta}\supf[]\expect{\boldx,\boldy}{f(\boldx,\boldy)}-\expect{\gtheta,\boldy}{f(\gtheta,\boldy)}.
\end{gathered}
\end{equation}
If $f$ is a parametrized critic, this is a game between two adversaries, having a classic GAN structure.
While this optimization task may seem appealing and practical, it is in fact ill-posed since we are confined to a finite sized and unbalanced data-set, in which for each $\boldx$ we have many $\boldy's$ but not vice versa. 
A generator optimized under (\ref{obj:cgan}) with such data would try to learn to sample from the posterior distribution $\Pxgiveny$ with only one sample from
$\boldx|\boldy$ for each $\boldy$.
This would most likely lead to mode collapse~\cite{gan,pix2pix2017,mathieu2016deep,dsganICLR2019}, where $\gtheta|\boldy$ becomes a degenerate random variable and the generator ignores $\boldz$, since for each conditional input $\boldy$ it is sufficient for the generator to produce only one image that is acceptable by the critic. 
Hence, the densities $\Pggiveny$ and $\Pxgiveny$ might be equal only on this finite number of points, while allowing a deviation in the remaining domain.
To alleviate this weakness, we add a constraint to (\ref{obj:cgan}) as follows:
\begin{equation}
\begin{gathered}\label{obj:maintheorem_loss_expected}
\min_{\theta}\loss\\ 
s.t.\ \mathbb{E}_{\boldx,\boldy}\left[\norm{\boldx-\mathbb{E}\left[\mathbf{g}_{\theta}|\boldy\right]}_{2}^{2}\right]=\mathbb{E}_{\boldy}\left[Var(\boldx|\boldy)\right].
\end{gathered}
\end{equation}
Observe that if there exists $\theta^{*}$ such that $p_{\mathbf{g_{\theta^{*}}|\boldy}}=p_{\mathbf{\boldx|\boldy}}$, then $\theta^{*}$ is a global optimum of task (\ref{obj:cgan}), implying that the distance between the two conditional distributions is zero.
In addition, $\theta^{*}$ remains the global optimum of task (\ref{obj:maintheorem_loss_expected}) since $p_{\mathbf{g_{\theta^{*}}|\boldy}}=p_{\mathbf{\boldx|\boldy}}$ implies that $\mathbb{E}[\mathbf{g_{\theta^{*}}}|\boldy]=\mathbb{E}[\boldx|\boldy]$, and thus $\mathbb{E}_{\boldx,\boldy}\left[\norm{\boldx-\mathbb{E}\left[\mathbf{g}_{\theta^{*}}|\boldy\right]}_{2}^{2}\right]=\mathbb{E}_{\boldx,\boldy}\left[\norm{\boldx-\mathbb{E}\left[\boldx|\boldy\right]}_{2}^{2}\right]=\expect{\boldy}{Var(\boldx|\boldy)}$, which is the Minimum Mean Squared Error (MMSE).
Thus, the added constraint is also satisfied by $\theta^{*}$, which means that it is a necessary condition on any solution that yields $p_{\mathbf{g_{\theta^{*}}|\boldy}}=p_{\mathbf{\boldx|\boldy}}$.
In other words, instead of having distortion requirements on specific samples (as in LAG~\cite{berthelot2020creating} for instance), we require an agreement with the expectation of the posterior, i.e., the constraint enforces many samples of $\gtheta|\boldy$ to agree with $\boldx|\boldy$ (in expectation).
As we will see in \autoref{experiments}, this revision leads to a stochastic variation and therefore circumvents mode collapse.

Since $\expect{\boldy}{Var(\boldx|\boldy)}$ is the global minimum of $\mathbb{E}_{\boldx,\boldy}\left[\norm{\boldx-\mathbb{E}\left[\gtheta|\boldy\right]}_{2}^{2}\right]$, we can reformulate the optimization of task (\ref{obj:maintheorem_loss_expected}) by adding a penalty term to task (\ref{obj:cgan}):
\begin{equation}\label{obj:final_opt}
\begin{gathered}
\min_{\theta}\loss+\lambda\mathbb{E}_{\boldx,\boldy}\left[\norm{\boldx-\mathbb{E}\left[\mathbf{g}_{\theta}|\boldy\right]}_{2}^{2}\right].
\end{gathered}
\end{equation}
The posterior distribution is still a globally optimal solution to this problem, since both expressions admit their minimum for the same $\theta^{*}$. This way, the proposed scheme eliminates many possible solutions that might minimize the first term but are far from the true posterior.

\begin{algorithm*}[h!]
\SetInd{1em}{1em}
\textbf{Require:} The gradient penalty coefficient $\lambda_{GP}$, the expected distance coefficient $\lambda_{MM}$, the number of critic iterations per generator iteration $n_{critic}$, the batch size $B$, the number of sampled realizations from the generator $M$, the penalty batch size $PB$, Adam hyperparameters $\alpha,\beta_{1},\beta_{2}$, initial critic and generator parameters $\omega_{0}$ and $\theta_{0}$.\\
\textbf{Default Settings:} $\lambda_{GP}=10,\lambda_{MM}=10^{-3},n_{critic}=1,B=32,M=8,PB=8,\alpha=2.5\cdot 10^{-4},\beta_{1}=0,\beta_{2}=0.99$.\\
\While{\text{$\theta$ has not converged}}{
  \For{$t=1,\hdots,n_{critic}$}{
    \For{$i=1,\hdots,B$}{
        Sample $x\sim\Px$, $z\sim\Pz$, $\epsilon\sim U[0,1]$\\
        $y\leftarrow\mathbf{deg}(x)$\ \tcp{A stochastic degradation operator.}
        $\tilde{x}\leftarrow G_{\theta}(z,y)$\\
        $\hat{x}\leftarrow\epsilon x+(1-\epsilon)\tilde{x}$\\
        $L_{C}^{(i)}\leftarrow \lambda_{MM}(C_{\omega}(\tilde{x},y)-C_{\omega}(x,y))+\lambda_{GP}(\norm{\nabla_{\hat{x}}C_{\omega}(\hat{x},y)}_{2}-1)^{2}$
    }
    $\omega\leftarrow \text{Adam}(\nabla_{\omega}\frac{1}{B}\sum_{i=1}^{B}L_{C}^{(i)},\omega,\alpha,\beta_{1},\beta_{2})$
  }
  \For{$i=1,\hdots,B$}{
    Sample $x\sim\Px$, $z^{(0)}\sim\Pz$\\
    $y\leftarrow\mathbf{deg}(x)$\\
    $L_{G_{MM}}^{(i)}\leftarrow-\lambda_{MM}C_{\omega}(G_{\theta}(z^{(0)},y),y)$\\
    \If{$i\leq PB$}{
        Sample a batch $\{z^{(j)}\}_{j=1}^{M}$, each from $\Pz$\\
        $L_{G_{A}}^{(i)}\leftarrow\norm{x-\frac{1}{M}\sum_{j=1}^{M}G_{\theta}(z^{(j)},y)}_{2}^{2}$
    }
  }
  $\theta\leftarrow \text{Adam}(\nabla_{\theta}\left[\frac{1}{B}\sum_{i=1}^{B}L_{G_{MM}}^{(i)}+\frac{1}{PB}\sum_{i=1}^{PB}L_{G_{A}}^{(i)}\right],\theta,\alpha,\beta_{1},\beta_{2})$
}
\caption{Training of the Posterior Sampling CGAN (PSCGAN).}
\label{alg:ours}
\end{algorithm*}
At first glance, LAG~\cite{berthelot2020creating} could also seem like a method that directly aims for the perceptual quality goal.
LAG's objective is almost identical to that of CGAN, with an additional generator regularization term:
\begin{equation}\label{obj:lag_opt}
\begin{gathered}
\min_{\theta}\loss\\
+\lambda\mathbb{E}_{\boldx,\boldy}\left[\norm{P(\boldx,\boldy)-P(G_{\theta}(0,\boldy),\boldy)}_{2}^{2}\right].
\end{gathered}
\end{equation}
In LAG, the function $P(\cdot,\cdot)$ in the above expression represents a part of the critic that extracts features from the image pair fed to it.
This function could be considered as a variant or part of the function $f$ we have used above.
Referring to the second term, its rationale is the belief that the intermediate representations of $(\boldx,\boldy)$ and $(G_{\theta}(0,\boldy),\boldy)$ are necessarily close-by.
Observe that this penalty is substantially different from the expectation requirement we have posed in equation (\ref{obj:final_opt}), since we do not assume that a given sample (i.e., the one attained at $\boldz=0$) and $\boldx$ are matched in distortion.
When $G_{\theta}(\cdot,\cdot)$ and $P(\cdot,\cdot)$ are continuous mappings, 
this assumption poses a distortion requirement not only on $G_{\theta}(0,\boldy)$, but also on its neighbourhood.
Thus, the penalty term in (\ref{obj:lag_opt}) conflicts with the perceptual quality goal~\cite{Blau_2018}, whereas the penalty term we propose in (\ref{obj:final_opt}) does not.
\section{Proposed Method: Details} \label{proposed}
\subsection{Training Method}
Our training method is directly derived from optimization task (\ref{obj:final_opt}).
To enforce the 1-Lipschitz constraint on the critic (denoted as $f$ in equation (\ref{obj:cgan})), we use the gradient penalty version of WGAN~\cite{gulrajani2017improved}.
That is, we train a generator $G_{\theta}$ and a critic $C_{\omega}$ (replacing $f$, to align with common WGAN notations) via the min-max optimization game
\begin{align}\label{obj:training_method}
\min_{\theta}\max_{\omega}\  
&\mathbb{E}_{\boldx,\boldy}\left[\norm{\boldx-\mathbb{E}_{\boldz}\left[G_{\theta}(\boldz,\boldy)|\boldy\right]}_{2}^{2}\right]\\
+\lambda_{MM}&\expect{\boldx,\boldy}{C_{\omega}(\boldx,\boldy)}-\expect{\boldz,\boldy}{C_{\omega}(G_{\theta}(\boldz,\boldy),\boldy)}\nonumber\\
+\lambda_{GP}\ &\mathbb{E}_{\hat{\boldx},\boldy}\left[(\norm{\nabla_{\mathbf{\hat{x}}}C_{\omega}(\mathbf{\hat{x}},\boldy)}_{2}-1)^{2}\right]\nonumber,
\end{align}
where for a given $\boldy=y$, the last expectation is taken with respect to $\mathbb{P}_{\mathbf{\hat{x}}}$, the distribution of uniform samples along straight lines between pairs of points sampled from $\mathbb{P}_{\boldx|\boldy=y}$ and $\mathbb{P}_{\mathbf{g_{\theta}}|\boldy=y}$.
Our proposed training method is described in \autoref{alg:ours}, and our proposed generator architecture and a full framework schematic are disclosed in \autoref{generator_architecture}.
\begingroup
\setlength{\tabcolsep}{0pt} 
\renewcommand{\arraystretch}{0} 
    \begin{figure*}[t]
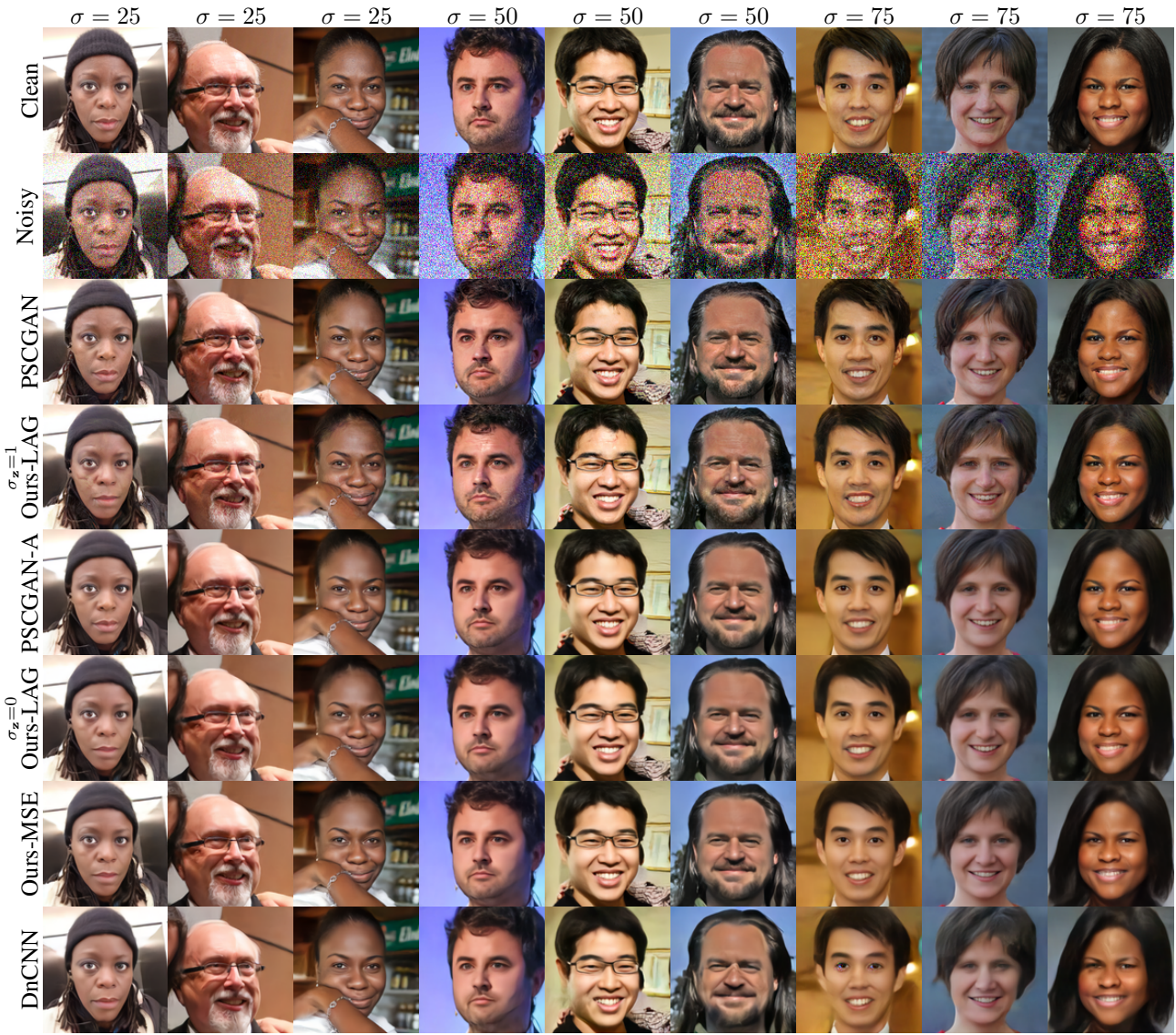

        \centering
        \begin{tabular}{r c c c c c c c c c}
        &$\sigma=25$&$\sigma=25$&$\sigma=25$&$\sigma=50$&$\sigma=50$&$\sigma=50$&$\sigma=75$&$\sigma=75$&$\sigma=75$\\
        \rule{0pt}{0.4ex}&&&&&&&&&\\
        \centered{\begin{turn}{90}Clean\end{turn}}\rule{2pt}{0ex}&
        \real{25}{ffhq}{3}{1} &
        \real{25}{ffhq}{4}{4} &
        \real{25}{ffhq}{5}{0} &
        \real{50}{ffhq}{25}{6}&
        \real{50}{ffhq}{26}{6}&
        \real{50}{ffhq}{23}{7}&
        \real{75}{ffhq}{0}{7} &
        \real{75}{ffhq}{14}{1}&
        \real{75}{ffhq}{25}{5}\\
        \centered{\begin{turn}{90}Noisy\end{turn}}\rule{2pt}{0ex}&
        \noisy{25}{ffhq}{3}{1} &
        \noisy{25}{ffhq}{4}{4} &
        \noisy{25}{ffhq}{5}{0} &
        \noisy{50}{ffhq}{25}{6}&
        \noisy{50}{ffhq}{26}{6}&
        \noisy{50}{ffhq}{23}{7}&
        \noisy{75}{ffhq}{0}{7} &
        \noisy{75}{ffhq}{14}{1}&
        \noisy{75}{ffhq}{25}{5}\\
        \centered{\begin{turn}{90}PSCGAN\end{turn}}\rule{2pt}{0ex}&
        \ourss{25}{ffhq}{3}{1} {00}&
        \ourss{25}{ffhq}{4}{4} {00}&
        \ourss{25}{ffhq}{5}{0} {00}&
        \ourss{50}{ffhq}{25}{6}{00}&
        \ourss{50}{ffhq}{26}{6}{00}&
        \ourss{50}{ffhq}{23}{7}{00}&
        \ourss{75}{ffhq}{0}{7} {00}&
        \ourss{75}{ffhq}{14}{1}{00}&
        \ourss{75}{ffhq}{25}{5}{00}\\
        \centered{\begin{turn}{90}$\overset{{\tiny\sigma_{\boldz}=1}}{\text{Ours-LAG}}$\end{turn}}\rule{2pt}{0ex}&
        \lag{25}{ffhq}{3}{1} {02}&
        \lag{25}{ffhq}{4}{4} {00}&
        \lag{25}{ffhq}{5}{0} {00}&
        \lag{50}{ffhq}{25}{6}{00}&
        \lag{50}{ffhq}{26}{6}{00}&
        \lag{50}{ffhq}{23}{7}{00}&
        \lag{75}{ffhq}{0}{7} {00}&
        \lag{75}{ffhq}{14}{1}{00}&
        \lag{75}{ffhq}{25}{5}{00}\\
        \centered{\begin{turn}{90}PSCGAN-A\end{turn}}\rule{2pt}{0ex}&
        \oursa{25}{ffhq}{3}{1} &
        \oursa{25}{ffhq}{4}{4} &
        \oursa{25}{ffhq}{5}{0} &
        \oursa{50}{ffhq}{25}{6}&
        \oursa{50}{ffhq}{26}{6}&
        \oursa{50}{ffhq}{23}{7}&
        \oursa{75}{ffhq}{0}{7} &
        \oursa{75}{ffhq}{14}{1}&
        \oursa{75}{ffhq}{25}{5}\\
        \centered{\begin{turn}{90}$\overset{{\tiny\sigma_{\boldz}=0}}{\text{Ours-LAG}}$\end{turn}}\rule{2pt}{0ex}&
        \laglikely{25}{ffhq}{3}{1} &
        \laglikely{25}{ffhq}{4}{4} &
        \laglikely{25}{ffhq}{5}{0} &
        \laglikely{50}{ffhq}{25}{6}&
        \laglikely{50}{ffhq}{26}{6}&
        \laglikely{50}{ffhq}{23}{7}&
        \laglikely{75}{ffhq}{0}{7} &
        \laglikely{75}{ffhq}{14}{1}&
        \laglikely{75}{ffhq}{25}{5}\\
        \centered{\begin{turn}{90}Ours-MSE\end{turn}}\rule{2pt}{0ex}&
        \oursmse{25-ffhq}{3}&
        \oursmse{25-ffhq}{4}&
        \oursmse{25-ffhq}{5}&
        \oursmse{50-ffhq}{25}&
        \oursmse{50-ffhq}{26}&
        \oursmse{50-ffhq}{23}&
        \oursmse{75-ffhq}{0} &
        \oursmse{75-ffhq}{14}&
        \oursmse{75-ffhq}{25}\\
        \centered{\begin{turn}{90}DnCNN\end{turn}}\rule{2pt}{0ex}&
        \dncnn{25}{3}&
        \dncnn{25}{4}&
        \dncnn{25}{5}&
        \dncnn{50}{25}&
        \dncnn{50}{26}&
        \dncnn{50}{23}&
        \dncnn{75}{0} &
        \dncnn{75}{14}&
        \dncnn{75}{25}
        \end{tabular}
        \caption{Denoising results on the FFHQ test set produced by several methods. 
        PSCGAN is a sampled denoised image produced by our proposed method, attained by injecting noise with standard deviation of $\sigma_{\boldz}=1$ (at both training and inference time).
       $\text{Ours-LAG}(\tiny\sigma_{\boldz}=0)$ and $\text{Ours-LAG}(\tiny\sigma_{\boldz}=1)$ are the same models, while the former is with $\sigma_{\boldz}=0$ and the latter is with $\sigma_{\boldz}=1$ at inference time.
        In this case, PSCGAN-A averages 64 instances of PSCGAN. 
        Each model was trained on the FFHQ training set to denoise a specific noise level ($25, 50$ or $75$).}
        \label{fig:collage}
    \end{figure*}
\endgroup

\begingroup
\newcolumntype{M}[1]{>{\centering\arraybackslash}m{#1}}
\setlength{\tabcolsep}{0pt} 
\renewcommand{\arraystretch}{0} 
    \begin{figure*}[t!]
     \centering
     \begin{tabular}{c M{0.05\textwidth}}
        \begin{tabular}{c p{0.025\textwidth} c c c c c c}
        &&&\multicolumn{4}{c}{Denoised Samples}&\\
        &&Noisy&\multicolumn{4}{c}{\downbracefill}&Std Dev\\
        &\rule{0pt}{0.8ex}&&&&&\\
        \begin{tabular}{c}
             \vspace{1.3cm}\\
             Clean
        \end{tabular}&\hspace{3pt}\centered{\begin{turn}{90}$\sigma=25$\end{turn}}&
        \noisy{25}{ffhq}{15}{2}&
        \ourss{25}{ffhq}{15}{2}{01}&
        \ourss{25}{ffhq}{15}{2}{00}&
        \ourss{25}{ffhq}{15}{2}{02}&
        \ourss{25}{ffhq}{15}{2}{03}&
        \stddev{25}{ffhq}{15}{2}{1}\\
        \real{25}{ffhq}{15}{2}&
        \hspace{3pt}\centered{\begin{turn}{90}$\sigma=50$\end{turn}}&
        \noisy{50}{ffhq}{15}{2}&
        \ourss{50}{ffhq}{15}{2}{01}&
        \ourss{50}{ffhq}{15}{2}{00}&
        \ourss{50}{ffhq}{15}{2}{02}&
        \ourss{50}{ffhq}{15}{2}{03}&
        \stddev{50}{ffhq}{15}{2}{1}\\
        &\hspace{3pt}\centered{\begin{turn}{90}$\sigma=75$\end{turn}}&
        \noisy{75}{ffhq}{15}{2}&
        \ourss{75}{ffhq}{15}{2}{03}&
        \ourss{75}{ffhq}{15}{2}{02}&
        \ourss{75}{ffhq}{15}{2}{04}&
        \ourss{75}{ffhq}{15}{2}{05}&
        \stddev{75}{ffhq}{15}{2}{1}\\
        \end{tabular}&\vspace{1cm}\multirow{2}{*}{
        \begin{tikzpicture}
            \begin{axis}[
                hide axis,
                scale only axis,
                height=5cm,
                width=1cm,
                colormap = {whiteblack}{color(0cm)  = (white);color(1cm) = (black)},
                colorbar,
                point meta min=0,
                point meta max=1,
                colorbar style={
                }]
                \addplot [draw=none] coordinates {(0,0)};
            \end{axis}
            \end{tikzpicture}}\\
        \rule{0pt}{0.4ex}&\\
        \begin{tabular}{c p{0.025\textwidth} c c c c c c}
        \begin{tabular}{c}
             \vspace{1.3cm}\\
             Clean
        \end{tabular}&\hspace{3pt}\centered{\begin{turn}{90}$\sigma=25$\end{turn}}&
        \noisy{25}{ffhq}{15}{6}&
        \ourss{25}{ffhq}{15}{6}{01}&
        \ourss{25}{ffhq}{15}{6}{00}&
        \ourss{25}{ffhq}{15}{6}{02}&
        \ourss{25}{ffhq}{15}{6}{03}&
        \stddev{25}{ffhq}{15}{6}{1}\\
        \real{25}{ffhq}{15}{6}&
        \hspace{3pt}\centered{\begin{turn}{90}$\sigma=50$\end{turn}}&
        \noisy{50}{ffhq}{15}{6}&
        \ourss{50}{ffhq}{15}{6}{01}&
        \ourss{50}{ffhq}{15}{6}{00}&
        \ourss{50}{ffhq}{15}{6}{02}&
        \ourss{50}{ffhq}{15}{6}{03}&
        \stddev{50}{ffhq}{15}{6}{1}\\
        &\hspace{3pt}\centered{\begin{turn}{90}$\sigma=75$\end{turn}}&
        \noisy{75}{ffhq}{15}{6}&
        \ourss{75}{ffhq}{15}{6}{00}&
        \ourss{75}{ffhq}{15}{6}{02}&
        \ourss{75}{ffhq}{15}{6}{04}&
        \ourss{75}{ffhq}{15}{6}{05}&
        \stddev{75}{ffhq}{15}{6}{1}\\
        \end{tabular}&
\end{tabular}
        \caption{Stochastic variation of denoised images attained by 3 different generators, each trained with PSCGAN to denoise images contaminated with noise levels of $\sigma=25,50,75$.
        Two clean images are presented to the left, and their corresponding noisy versions to their right.
        Alongside each noisy input we show 4 examples of possible denoising outcomes, as well as the $4^{\text{th}}$ root of the per-pixel standard deviation image calculated on 32 samples.
        For convenience, a gray-scale color map is added to the right (white and black correspond to low and high standard deviations, respectively).
        All denoised image samples were obtained by injecting noise with $\sigma_{\boldz}=1$ at inference time.}
        \label{fig:stochastic_variation}
    \end{figure*}
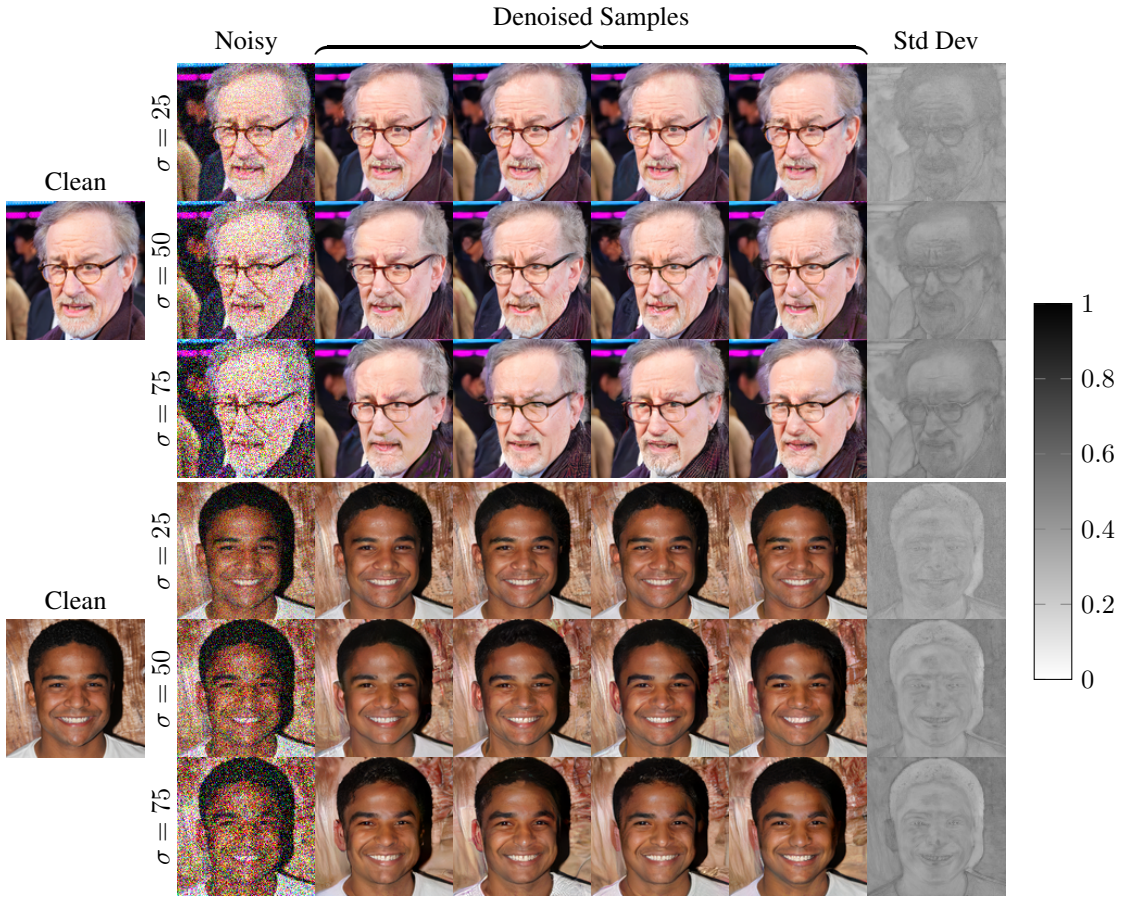
\endgroup
\subsection{A Denoiser with Two Distinct Capabilities}\label{two_for_one}
Recall that in our training method we drive our generator towards the production of samples from the posterior distribution while constraining the average denoised image to minimize the MSE.
Thus, given such a trained model with enough capacity, the average denoised image of a given noisy input should approximately achieve the MMSE.
This allows the optimized model to go beyond sampling from the posterior, producing an MMSE approximation result by averaging many generated samples.
Even if the trained model does not accurately capture the true posterior, an average denoised image should produce low MSE, while a sampled denoised image should produce high perceptual quality.
Our training method therefore allows one to obtain two denoisers at the same time: a denoiser that approximately samples from the posterior distribution and achieves high perceptual quality, and a denoiser that approximates the MMSE estimator.
In \autoref{experiments} we refer to the former as \emph{PSCGAN} and to the latter as \emph{PSCGAN-A}.
\section{Experimental Evaluation}  \label{experiments}
\noindent We turn to present an evaluation of several methods:
\begin{itemize}[leftmargin=*]
    \item PSCGAN, our proposed method, which aims to sample from the posterior distribution.
    \item PSCGAN-A, which averages instances of PSCGAN.
    \item Ours-MSE, which is our proposed generator trained to solely optimize the MSE loss (without noise injections). Comparing its performance with PSCGAN allows us to better evaluate our proposed training method to an MSE based optimization procedure since we use the same architecture in both cases.
    \item DnCNN~\cite{Zhang_DnCnn_2017}, a commonly accepted baseline.
\end{itemize}
Additionally, we evaluate \emph{Ours-LAG}, a variant of LAG~\cite{berthelot2020creating}, only to illustrate several tendencies regarding the perception-distortion tradeoff in \autoref{perception_distortion_tradeoff_section}.
In \autoref{lag_implementation} we describe our implementation choices.

In all experiments of PSCGAN the noise injected to the generator is of Gaussian distribution with zero mean.
We vary at inference time the standard deviation of all noisy maps injected to the generator, which we denote as $\sigma_{\boldz}$ ($\sigma_{\boldz}=0$ means $\boldz=0$).
To clarify, PSCGAN is always trained with $\sigma_{\boldz}=1$.
We also vary the number of instances produced by PSCGAN that are being averaged to compute PSCGAN-A, which we denote by $N$.
We base our evaluations on the FFHQ~\cite{Karras_2019_CVPR} thumbnails, LSUN Bedroom and LSUN Church outdoor~\cite{LSUNdataset} data sets, and assess the performance on images contaminated with different levels of additive white Gaussian noise with $\sigma\in\{25, 50, 75\}$. 
To clarify, we train a separate denoiser for each configuration of data set and noise level.
Supplementary training details are in \autoref{training}.
\subsection{Perceptual Quality and Distortion Evaluation}
\setlength{\tabcolsep}{0.35em}
\begin{table*}[t]
    \centering
    \small
    \begin{tabular}{|c|c|c|c|c|c|c|c|c|c|}
        \hline
        \multirow{2}{*}{\textbf{Data set}}& 
        \multirow{2}{*}{\textbf{$\sigma$}}&
        \multicolumn{2}{c|}{\textbf{PSCGAN}}&
        \multicolumn{2}{c|}{\textbf{PSCGAN-A}}&
        \multicolumn{2}{c|}{\textbf{Ours-MSE}}&
        \multicolumn{2}{c|}{\textbf{DnCNN}}\\
        \cline{3-10}&&PSNR&FID&PSNR&FID&PSNR&FID&PSNR&FID\\
        \hline
        \multirow{3}{*}{\textbf{FFHQ}} & 25 & 29.19&$\textbf{12.66}\pm 0.07$ & 31.46&27.48 &\textbf{31.83}&31.48& 31.77&36.80\\
        \cline{2-10}& 50 & 25.83&$\textbf{15.18}\pm 0.15$ & 28.28&31.81 & \textbf{28.44}&41.56& 28.30&42.97\\
        \cline{2-10}& 75& 24.09&$\textbf{15.78} \pm 0.13$& 26.57&34.64 & \textbf{26.81}&46.31& 26.46&47.69\\
        \hline
        \multirow{3}{*}{\textbf{\begin{tabular}{c}
             LSUN\\Church
        \end{tabular}}} & 25 &
        29.03&
        $\textbf{7.66}\pm 0.04$ &
        30.78&
        9.33&
        \textbf{31.20}&
        9.69&
        31.16&
        10.25\\
        \cline{2-10}& 50 &
        25.50& 
        $\textbf{9.02}\pm 0.06$ &
        27.54&
        10.86&
        \textbf{27.77} &
        12.93&
        27.69&
        15.66\\
        \cline{2-10}& 75 &
        23.75 &
        $\textbf{9.12}\pm 0.09$ &
        25.84&
        12.39&
        \textbf{26.00} &
        14.94&
        25.78&
        22.12\\
        \hline
        \multirow{3}{*}{\textbf{\begin{tabular}{c}
             LSUN\\Bedroom
        \end{tabular}}} & 25 &
        30.62&
        $\textbf{8.83}\pm 0.05$ &
        32.29&
        9.41&
        \textbf{32.57}&
        11.86&
        32.02&
        11.32\\
        \cline{2-10}& 50 &
        27.30&
        $\textbf{9.27}\pm 0.06$&
        29.08&
        11.13&
        \textbf{29.30}& 
        12.71&
        29.10&
        21.38\\
        \cline{2-10}& 75 &
        25.23&
        $\textbf{11.56}\pm 0.08$ & 
        27.26&
        13.74&
        \textbf{27.43}&
        15.57&
        27.14&
        31.69\\
        \hline
    \end{tabular}
    \caption{The PSNR (dB) and FID results obtained by several evaluated methods, each trained to denoise images contaminated with a specific noise level (higher PSNR and lower FID correspond to better performance). Notice that the reported PSNR is not the average one, but rather the PSNR calculated on the average MSE of the entire test set. PSCGAN is our sampler from the learned distribution, where we use $\sigma_{\boldz}=1$ for the FFHQ test set and $\sigma_{\boldz}=0.75$ for both LSUN test sets during inference.
    In this case, $\text{PSCGAN-A}$ averages $N=64$ instances of PSCGAN (obtained with $\sigma_{\boldz}=1$ on all data sets). 
    Ours-MSE is our proposed generator trained to solely optimize the MSE loss (without noise injections).
    The FID reports of PSCGAN contain both the mean and the standard deviation (denoted with $\pm$).
    }
    \label{tab:psnr}
\end{table*}
\definecolor{palecarmine}{rgb}{0.69, 0.25, 0.21}
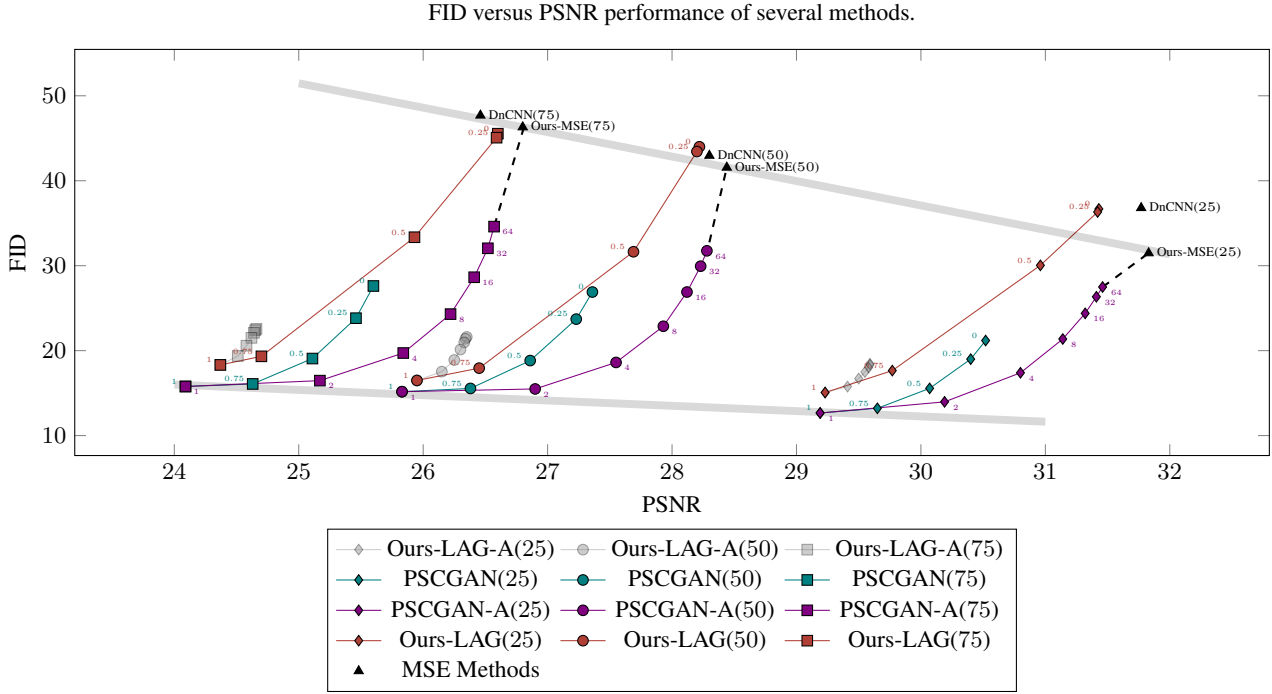
\begin{figure*}[t]
\centering
\small
\begin{tikzpicture}
\begin{axis} [
        title=FID versus PSNR performance of several methods.,
        xlabel=PSNR,
        ylabel=FID,
        width=1\textwidth,
        height=0.4\textwidth,
        nodes near coords,
        every node near coord/.append style={font=\tiny},
        point meta=explicit symbolic,
    legend style={
        at={(0.5,-0.18)},
        anchor=north,
        legend columns=3
    }]

\addplot+[opacity=0.4,solid, mark=diamond*, color=gray ,mark options={draw=black, fill=gray}, every node near coord/.append style={anchor=east,xshift=0pt,yshift=2pt}] table [x=psnr, y=fid] {figures/data/fid_psnr_lag-a_25.dat};
\addlegendentryexpanded{Ours-LAG-A($25$)}
\addplot+[opacity=0.4,solid, mark=*, color=gray ,mark options={draw=black, fill=gray}, every node near coord/.append style={anchor=east,xshift=0pt,yshift=2pt}] table [x=psnr, y=fid] {figures/data/fid_psnr_lag-a_50.dat};
\addlegendentryexpanded{Ours-LAG-A($50$)}
\addplot+[opacity=0.4,solid, mark=square*, color=gray ,mark options={draw=black, fill=gray}, every node near coord/.append style={anchor=east,xshift=0pt,yshift=2pt}] table [x=psnr, y=fid] {figures/data/fid_psnr_lag-a_75.dat};
\addlegendentryexpanded{Ours-LAG-A($75$)}

\addplot+[solid, mark=diamond*, color=teal, mark options={draw=black, fill=teal}, every node near coord/.append style={anchor=east,xshift=0pt,yshift=2pt}] table [x=psnr, y=fid, meta=sigma_z] {figures/data/fid_psnr_ours-s_25.dat};
\addlegendentryexpanded{PSCGAN($25$)}
\addplot+[solid, mark=*, color=teal, mark options={draw=black, fill=teal}, every node near coord/.append style={anchor=east,xshift=0pt,yshift=2pt}] table [x=psnr, y=fid, meta=sigma_z] {figures/data/fid_psnr_ours-s_50.dat};
\addlegendentryexpanded{PSCGAN($50$)}
\addplot+[solid, mark=square*, color=teal, mark options={draw=black, fill=teal}, every node near coord/.append style={anchor=east,xshift=0pt,yshift=2pt}] table [x=psnr, y=fid, meta=sigma_z] {figures/data/fid_psnr_ours-s_75.dat};
\addlegendentryexpanded{PSCGAN($75$)}

\addplot+[solid, mark=diamond*, color=violet ,mark options={draw=black, fill=violet},every node near coord/.append style={anchor=west,xshift=0pt,yshift=-2pt}] table [x=psnr, y=fid, meta=n] {figures/data/fid_psnr_ours-a_25.dat};
\addlegendentryexpanded{PSCGAN-A($25$)}
\addplot+[solid, mark=*, color=violet ,mark options={draw=black, fill=violet},every node near coord/.append style={anchor=west,xshift=0pt,yshift=-2pt}] table [x=psnr, y=fid, meta=n] {figures/data/fid_psnr_ours-a_50.dat};
\addlegendentryexpanded{PSCGAN-A($50$)}
\addplot+[solid, mark=square*, color=violet ,mark options={draw=black, fill=violet}, every node near coord/.append style={anchor=west,xshift=0pt,yshift=-2pt}] table [x=psnr, y=fid, meta=n] {figures/data/fid_psnr_ours-a_75.dat};
\addlegendentryexpanded{PSCGAN-A($75$)}

\addplot+[solid, mark=diamond*, color=palecarmine ,mark options={draw=black, fill=palecarmine},every node near coord/.append style={anchor=east,xshift=0pt,yshift=2pt}] table [x=psnr, y=fid, meta=sigma_z] {figures/data/fid_psnr_lag_25.dat};
\addlegendentryexpanded{Ours-LAG($25$)}
\addplot+[solid, mark=*, color=palecarmine ,mark options={draw=black, fill=palecarmine},every node near coord/.append style={anchor=east,xshift=0pt,yshift=2pt}] table [x=psnr, y=fid, meta=sigma_z] {figures/data/fid_psnr_lag_50.dat};
\addlegendentryexpanded{Ours-LAG($50$)}
\addplot+[solid, mark=square*, color=palecarmine ,mark options={draw=black, fill=palecarmine}, every node near coord/.append style={anchor=east,xshift=0pt,yshift=2pt}] table [x=psnr, y=fid, meta=sigma_z] {figures/data/fid_psnr_lag_75.dat};
\addlegendentryexpanded{Ours-LAG($75$)}

\addplot+[solid, mark=triangle*, mark options={draw=black, fill=black}, only marks, every node near coord/.append style={font=\tiny, color=black,anchor=west}] table [x=psnr, y=fid, meta=label] {figures/data/fid_psnr_mse_methods.dat};
\addlegendentryexpanded{MSE Methods}

\draw [thick, dashed] (26.81, 46.31) -- (26.57, 34.60);
\draw [thick, dashed] (31.46, 27.49) -- (31.83, 31.48);
\draw [thick, dashed] (28.30, 32.75) -- (28.44, 41.56);

\addplot[domain=25:32,  opacity=0.15, line width=3pt] {-2.877*x + 123.4};
\addplot[domain=24:31,  opacity=0.15, line width=3pt] {-0.6237*x + 30.97};
\end{axis}
\end{tikzpicture}
\caption{FID versus PSNR results for PSCGAN, PSCGAN-A that averages $N$ PSCGAN instances, Ours-LAG, Ours-LAG-A that averages $N$ Ours-LAG instances, Ours-MSE and DnCNN.
The noise contamination level ($\sigma=25,50,75$) is given with parentheses next to the name of each method.
PSCGAN and PSCGAN-A are evaluated on different choices of $\sigma_{\boldz}$ and $N$ during inference, while $\sigma_{\boldz}$ is fixed to 1 when varying $N$.
Ours-LAG and Ours-LAG-A are evaluated in the same fashion.
For PSCGAN and Ours-LAG the values of $\sigma_{\boldz}$ are given next to each marked point.
Similarly, the values of $N$ are given for PSCGAN-A.
The performance results of the MSE based methods are also plotted.}
\label{fig:perception_distortion_tradeoff}
\end{figure*}
In \autoref{fig:collage} we demonstrate the perceptual quality of all evaluated methods on the FFHQ test set, including Ours-LAG.
The visual results produced on both LSUN test sets are in \autoref{appendix:lsun_visual_results}.
PSCGAN and Ours-LAG (at $\sigma_{\boldz}=1$) produce sharp and real looking results and outperform the MSE methods in terms of perceptual quality, as the latter produce unnaturally smooth images.

Recall that PSCGAN is a stochastic denoiser able to produce many denoised outputs, and such variability is demonstrated in \autoref{fig:stochastic_variation}.
Even though the overall appearance of the varying samples is similar, we observe a rich stochastic variation on fine details such as wrinkles, hair, eyes and more.
In the same figure we also show the $4^{\text{th}}$ root of the per-pixel standard deviation calculated on 32 denoised samples of the same noisy input.
These results suggest that the penalty term in optimization objective (\ref{obj:final_opt}) indeed circumvents the aforementioned mode collapse issue.
In addition, it appears that our model does not suffer from inherent bias when handling skin tones, possibly due to the richness of the FFHQ data set~\cite{Karras_2019_CVPR}.

To quantitatively evaluate our proposed method we use the Fréchet Inception Distance (FID)~\cite{fid}, which is known to correlate well with human opinion scores.
Reliably computing FID requires a large amount of ``real'' samples (typically, at least 50,000), but evidently does not require many ``fake'' samples to remain consistent~\cite{mathiasen2020fast}.
To confirm this, we measure the FID (by using~\cite{Seitzer2020FID}) between each of the training sets and 100 randomly chosen subsets of 500 images taken from its corresponding test set, and see negligible variability in the scores.
Thus, our procedure to measure the FID of each denoising algorithm is to consider its outputs on each test set to its entirety as the fake samples, and use all the clean images of the corresponding training set as the real ones.
Since PSCGAN produces stochastic denoised images, we evaluate its average FID  by repeating this procedure 32 times, where in each  the FID is calculated by producing one realization of denoised image for each noisy input.

We report in \autoref{tab:psnr} the PSNR and FID scores obtained by all evaluated methods (except for Ours-LAG, which we evaluate only on the FFHQ test set in \autoref{perception_distortion_tradeoff_section}).
Several tendencies should be highlighted: 
\begin{itemize}[leftmargin=*]
\item As expected, in terms of PSNR, the best performing methods are the MMSE ones, with a gap of less than 3dB between these and PSCGAN, just as anticipated in~\cite{Blau_2018}. 
\item PSCGAN-A that averages 64 PSCGAN instances provides a very good approximation for the MMSE denoiser.
\item In FID terms, PSCGAN outperforms all other methods in all configurations, achieving superior perceptual quality.
\end{itemize}
It is important to note that the reported FID results are not necessarily the optimal ones, and thus we do not claim optimality.
We use this measure to provide some quantitative evaluation of the perceptual quality and to illustrate the perception-distortion tradeoff~\cite{Blau_2018} in the next section, but do not perform hyperparameter tuning to achieve the best FID.
\subsection{Traversing The Perception-Distortion Tradeoff}\label{perception_distortion_tradeoff_section}
Both PSCGAN and Ours-LAG allow traversing the perception-distortion tradeoff in two ways: by varying $\sigma_{\boldz}$ or by varying $N$.
We vary $\sigma_{\boldz}$ with values taken from $\{0,0.25,0.5,0.75,1\}$, and vary $N$ with values taken from $\{1,2,4,8,16,32,64\}$ (while fixing $\sigma_{\boldz}=1$).
We demonstrate the above traversals on the FFHQ test set in \autoref{fig:perception_distortion_tradeoff}, along with the FID and PSNR scores obtained by all evaluated MSE based methods.
For PSCGAN and Ours-LAG we report the average FID scores and omit their standard deviations since they are negligible.
We observe that:
\begin{itemize}[leftmargin=*]
    \item As theoretically expected, PSCGAN-A approaches Ours-MSE as $N$ increases, which suggests that our training procedure was successful in satisfying the penalty term, since with the same architecture we see a comparable PSNR performance when solely optimizing the MSE.
    \item For PSCGAN, varying $N$ (while fixing $\sigma_{\boldz}=1$) is more effective than varying $\sigma_{\boldz}$ (while fixing $N=1$), since each choice of $\sigma_{\boldz}$ is \emph{dominated}~\cite{Blau_2018} by some choice of $N$.
    In contrast, varying $\sigma_{\boldz}$ is more effective for Ours-LAG.
    We leave the explanation of these for future research.
    \item The FID performance of PSCGAN is only slightly affected by the noise level, suggesting that PSCGAN leads to high perceptual quality regardless of the noise contamination severity.
    This aligns with the posterior sampler's property to always produce images with perfect perceptual quality~\cite{Blau_2018}.
    In contrast, the PSNR performance of PSCGAN decreases as the noise level increases, which makes the perception-distortion tradeoff more significant in higher noise levels (emphasized by the two linear lines that diverge as the noise level increases).
    This evidence suggests that the gap in the perceptual quality of images produced by the posterior sampler and the MMSE estimator does not remain constant with the noise level, unlike the constant 3dB gap in PSNR~\cite{Blau_2018}.
    \item Averaging instances of Ours-LAG leads to a mild effect on the FID and PSNR scores. We leave the explanation of this phenomenon for future research.
    \item The results of Ours-LAG at $\sigma_{\boldz}=0$ and $0.25$ are almost identical, emphasizing that the low distortion requirement on $G_{\theta}(\boldz=0,\boldy)$ constrains its neighborhood with a similar requirement, when $G(\cdot,\cdot)$ is a continuous mapping.
    Indeed, when $\sigma_{\boldz}=1$, even though both Ours-LAG and PSCGAN use the same generator and critic architectures, the former slightly outperforms the latter in PSNR, while the opposite is true in FID.
    As claimed in \autoref{theoretical}, this shows a conflict with the perceptual quality goal~\cite{Blau_2018} at $\sigma_{\boldz}=1$.
    Consequently, we hypothesize that this leads PSCGAN-A to dominate Ours-LAG, since the latter finds a middle ground between the perceptual quality at $\sigma_{\boldz}=1$ and the distortion performance at $\sigma_{\boldz}=0$.
    \item The traversal curves of Ours-LAG are more ``stretched'' than those of PSCGAN (when varying $\sigma_{\boldz}$).
    Both ``pull'' the $\sigma_{\boldz}=1$ points towards high perceptual quality (and therefore towards high distortion~\cite{Blau_2018}), while only Ours-LAG ``pulls'' the $\sigma_{\boldz}=0$ points towards low distortion (and therefore towards low perceptual quality~\cite{Blau_2018}).
\end{itemize}
\begin{figure*}[h!]
\centering
\begin{tikzpicture}
    \begin{groupplot}[xlabel=RMSE,
    group style = {group size = 3 by 1, horizontal sep = 50pt}, width = 5.6cm, height = 3.9cm]
        \nextgroupplot[title = {$\sigma=25$},
         x tick label style={
        /pgf/number format/.cd,
            fixed,
            fixed zerofill,
            precision=2,
        /tikz/.cd
    },
        ylabel=Probability Density,
            legend style = {column sep = 10pt, legend columns = 2, legend to name = grouplegend}]

                  \addplot[line width=1.1, smooth, color=teal, no marks] table[x index=1,y index=0, each nth point={20}] {figures/data/hist25_PSCGAN.dat};
                  \addlegendentry[black]{PSCGAN patch-RMSE}
                  \addplot[line width=1.1, smooth, color=black, no marks] table[x index=1,y index=0, each nth point={20}] {figures/data/hist25_ours-MSE.dat};
                  \addlegendentry[black]{Ours-MSE patch-RMSE}
                  \addplot[line width=2.1, smooth, color=gray, no marks] table[x index=1,y index=0, each nth point={20}] {figures/data/hist25_est_noise_PSCGAN.dat};
                  \addlegendentry[black]{PSCGAN local-remainder-noise-RMS}
                  \addplot[line width=1.1, smooth, color=blue, no marks] table[x index=1,y index=0, each nth point={20}] {figures/data/hist25_noise.dat};
             \addlegendentry[black]{local-noise-RMS}
                   \addplot[line width=1.1, smooth, color=violet, no marks] table[x index=1,y index=0, each nth point={20}] {figures/data/hist25_est_noise_ours-MSE.dat};
                  \addlegendentry[black]{Ours-MSE local-remainder-noise-RMS}
        \nextgroupplot[title = {$\sigma=50$}, xlabel=RMSE,
         x tick label style={
        /pgf/number format/.cd,
            fixed,
            fixed zerofill,
            precision=2,
        /tikz/.cd
    }]
                  \addplot[line width=1.1, smooth, color=teal, no marks] table[x index=1,y index=0, each nth point={20}] {figures/data/hist50_PSCGAN.dat};
      \addplot[line width=1.1, smooth, color=black, no marks, each nth point={20}] table[x index=1,y index=0] {figures/data/hist50_ours-MSE.dat};
      \addplot[line width=2.1, smooth, color=gray, no marks, each nth point={20}] table[x index=1,y index=0] {figures/data/hist50_est_noise_PSCGAN.dat};
      \addplot[line width=1.1, smooth, color=blue, no marks, each nth point={20}] table[x index=1,y index=0] {figures/data/hist50_noise.dat};
      \addplot[line width=1.1, smooth, color=violet, no marks, each nth point={20}] table[x index=1,y index=0] {figures/data/hist50_est_noise_ours-MSE.dat};
        \nextgroupplot[title = {$\sigma=75$},
         x tick label style={
        /pgf/number format/.cd,
            fixed,
            fixed zerofill,
            precision=2,
        /tikz/.cd
    }] 
      \addplot[line width=1.1, smooth, color=teal, no marks, each nth point={20}] table[x index=1,y index=0] {figures/data/hist75_PSCGAN.dat};
      \addplot[line width=1.1, smooth, color=black, no marks, each nth point={20}] table[x index=1,y index=0] {figures/data/hist75_ours-MSE.dat};
      \addplot[line width=2.1, smooth, color=gray, no marks, each nth point={20}] table[x index=1,y index=0] {figures/data/hist75_est_noise_PSCGAN.dat};
      \addplot[line width=1.1, smooth, color=blue, no marks, each nth point={20}] table[x index=1,y index=0] {figures/data/hist75_noise.dat};
      \addplot[line width=1.1, smooth, color=violet, no marks, each nth point={20}] table[x index=1,y index=0] {figures/data/hist75_est_noise_ours-MSE.dat};
    \end{groupplot}
    \node at ($(group c2r1) + (0,-2.9cm)$) {\ref{grouplegend}}; 
\end{tikzpicture}
\caption{The approximated p.d.f of the patch-RMSE and of the local-remainder-noise-RMS obtained by PSCGAN and by Ours-MSE, and the approximated p.d.f of the local-noise-RMS obtained by Gaussian noise.}
\label{fig:residual_energy_histogram}
\end{figure*}
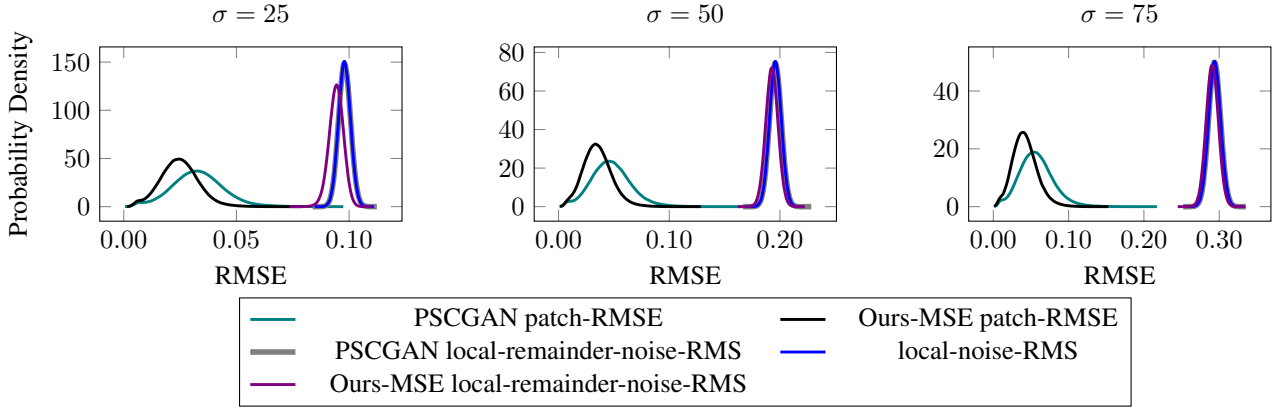
\subsection{Noise Reduction Evaluation}\label{noise_reduction_evaluation}
Image denoising is the process of recovering a clean signal $\boldx$ from a noisy observation $\boldy$, where in our case, $\boldy=\boldx+\boldn$ and $\boldn$ is white Gaussian noise.
This is an ill-posed inverse problem and usually $\boldx$ can not be fully retrieved.
Thus, a \emph{denoising algorithm} should find an approximation of $\boldx$, denoted as $\hat{\boldx}$, such that $\boldx$ and $\hat{\boldx}$ are close-by, and the \emph{remainder noise} $\hat{\boldn}=\boldy-\hat{\boldx}$ is normally distributed.
We assess whether PSCGAN, the algorithm which aims to sample from the posterior, satisfies such criteria.

The theoretical 3dB PSNR gap between the denoising results of a posterior sampler and of the MMSE estimator~\cite{Blau_2018} guarantees that a well trained model that aims to sample from the posterior distribution would produce denoised images that are close to their clean sources, when closeness is measured with MSE (as in many denoising algorithms).
Although the PSNR results obtained by PSCGAN (\autoref{tab:psnr}) are indeed high,
it clearly generates local content that is absent from the source image. Thus, we question whether the local patches of the reconstructed samples are still faithful to their clean counterparts.
We measure the \emph{patch-RMSE}, the square root of the MSE between all overlapping clean and denoised patches of size $15\times15$, for all images in the FFHQ test set and for both PSCGAN and Ours-MSE.
Finally, we create a histogram of the patch-RMSE values for each algorithm, and present the results in \autoref{fig:residual_energy_histogram}.
These give an approximation of the probability density function (p.d.f) of the patch-RMSE values obtained by each method.
In the same figure we also show the approximated p.d.f of the \emph{local-noise-RMS}, the root mean squared (RMS) value of all $15\times15$ patches of the noise $\boldn$ added to each clean image.
Likewise, we also show the approximated p.d.f of the \emph{local-remainder-noise-RMS}, referring to $\hat{\boldn}$.
Observe that the patch-RMSE obtained by Ours-MSE and by PSCGAN approximately follow the same p.d.f shape but with different mean and standard deviation.
Moreover, the p.d.fs of the local-noise-RMS and of the patch-RMSE obtained by PSCGAN are distant, the mean of the former being much larger.
Lastly, the p.d.f of the local-remainder-noise-RMS obtained by PSCGAN cannot be distinguished from that of the
local-noise-RMS, while the one obtained by Ours-MSE can, especially in lower noise levels.
These results suggest that noise elimination is attained by PSCGAN even locally, which means that it is stable in the sense that it generally does not produce improper local details.

Next, we question whether the remainder noise is normally distributed.
We use PSCGAN to denoise each image in the FFHQ test set and perform D’Agostino and Pearson’s normality test\footnote{A normally distributed
random variable should have a p-value greater than a threshold $\alpha$. 
We use $\alpha=0.05$, and in this case a realization of such a variable should pass the test with $95\%$ confidence.}~\cite{normality_test} on all of the resulting remainder noise images (2000 noise images, each of size $128\times 128$).
In addition, for each remainder noise image we extract randomly chosen $15\times 15$ patches and also patches that correspond to the largest patch-RMSE values (20 of each, for a total of $20\cdot 20\cdot 2000$ patches), and assess if they are normally distributed as well.
We find that PSCGAN successfully passes all tests in all configurations, with a p-value $>0.05$ with high confidence.
This shows that PSCGAN's remainder noise is normally distributed both locally and globally.
\section{Summary}\label{conclusion}
In this work we revisit the image denoising task and focus on producing visually pleasing images, as opposed to distortion based methods that target best PSNR.
Our strategy relies on the perceptual quality and distortion guarantees of posterior sampling, and a novel design of a CGAN to meet these needs.
We introduce a new constraint to the CGAN framework that alleviates its difficulty to train in the case of high dimensional distributions, where each input has only one corresponding source example.
We propose novel encoder-decoder denoiser architecture and training method, leading to denoised images with high perceptual quality and acceptable distortion.
\section{Acknowledgements}
We thank Bahjat Kawar for his contribution to the development of this paper.
We also thank Gennadi Zaidsher and the IT team of Technion's CS Department for providing the resources conducting this research.
{\small
\bibliographystyle{ieee_fullname}
\bibliography{egbib}

\begin{thebibliography}{10}\itemsep=-1pt

\bibitem{adler2018deep}
Jonas Adler and Ozan Öktem.
\newblock Deep bayesian inversion.
\newblock {\em arXiv preprint arXiv:1811.05910}, 2018.

\bibitem{wgan}
Martin Arjovsky, Soumith Chintala, and L{\'e}on Bottou.
\newblock Wasserstein generative adversarial networks.
\newblock In Doina Precup and Yee~Whye Teh, editors, {\em Proceedings of the
  34th International Conference on Machine Learning}, volume~70, pages
  214--223. PMLR, 2017.

\bibitem{Bahat_2020_CVPR}
Yuval Bahat and Tomer Michaeli.
\newblock Explorable super resolution.
\newblock In {\em Proceedings of the IEEE Conference on Computer Vision and
  Pattern Recognition}, June 2020.

\bibitem{berthelot2020creating}
David Berthelot, Peyman Milanfar, and Ian Goodfellow.
\newblock Creating high resolution images with a latent adversarial generator.
\newblock {\em arXiv preprint arXiv:2003.02365}, 2020.

\bibitem{Blau_2018}
Yochai Blau and Tomer Michaeli.
\newblock The perception-distortion tradeoff.
\newblock {\em Proceedings of the IEEE Conference on Computer Vision and
  Pattern Recognition}, June 2018.

\bibitem{normality_test}
Ralph D'Agostino and E.~S. Pearson.
\newblock Tests for departure from normality. empirical results for the
  distributions of $b2$ and $\sqrt{b1}$.
\newblock {\em Biometrika}, 60(3):613--622, 1973.

\bibitem{Dey2020}
Ratnadeep Dey, Debotosh Bhattacharjee, and Mita Nasipuri.
\newblock Image denoising using generative adversarial network.
\newblock In J.~K. Mandal and Soumen Banerjee, editors, {\em Intelligent
  Computing: Image Processing Based Applications}, volume 1157, pages 73--90.
  Springer Singapore, 2020.

\bibitem{Divakar_2017_CVPR_Workshops}
Nithish Divakar and R. Venkatesh~Babu.
\newblock Image denoising via cnns: an adversarial approach.
\newblock In {\em Proceedings of the IEEE Conference on Computer Vision and
  Pattern Recognition Workshops}, July 2017.

\bibitem{falcon2019pytorch}
WA Falcon et~al.
\newblock Pytorch lightning.
\newblock \url{ https://github.com/PyTorchLightning/pytorch-lightning}, 2019.

\bibitem{gan}
Ian Goodfellow et~al.
\newblock Generative adversarial nets.
\newblock In Z. Ghahramani, M. Welling, C. Cortes, N. Lawrence, and K.~Q.
  Weinberger, editors, {\em Advances in Neural Information Processing Systems},
  volume~27, pages 2672--2680. Curran Associates, Inc., 2014.

\bibitem{gulrajani2017improved}
Ishaan Gulrajani et~al.
\newblock Improved training of wasserstein gans.
\newblock In I. Guyon, U.~V. Luxburg, S. Bengio, H. Wallach, R. Fergus, S.
  Vishwanathan, and R. Garnett, editors, {\em Advances in Neural Information
  Processing Systems}, volume~30, pages 5767--5777. Curran Associates, Inc.,
  2017.

\bibitem{fid}
Martin Heusel, Hubert Ramsauer, Thomas Unterthiner, Bernhard Nessler, and Sepp
  Hochreiter.
\newblock Gans trained by a two time-scale update rule converge to a local nash
  equilibrium.
\newblock In {\em Proceedings of the 31st International Conference on Neural
  Information Processing Systems}, NIPS'17, page 6629–6640, Red Hook, NY,
  USA, 2017. Curran Associates Inc.

\bibitem{pix2pix2017}
Phillip Isola, Jun-Yan Zhu, Tinghui Zhou, and Alexei~A. Efros.
\newblock Image-to-image translation with conditional adversarial networks.
\newblock In {\em Proceedings of the IEEE Conference on Computer Vision and
  Pattern Recognition}, July 2017.

\bibitem{karras2018progressive}
Tero Karras, Timo Aila, Samuli Laine, and Jaakko Lehtinen.
\newblock Progressive growing of gans for improved quality, stability, and
  variation.
\newblock {\em arXiv preprint arXiv:1710.10196}, 2018.

\bibitem{Karras_2020_CVPR}
Tero Karras et~al.
\newblock Analyzing and improving the image quality of stylegan.
\newblock In {\em Proceedings of the IEEE Conference on Computer Vision and
  Pattern Recognition}, June 2020.

\bibitem{Karras_2019_CVPR}
Tero Karras, Samuli Laine, and Timo Aila.
\newblock A style-based generator architecture for generative adversarial
  networks.
\newblock In {\em Proceedings of the IEEE Conference on Computer Vision and
  Pattern Recognition}, June 2019.

\bibitem{kingma2017adam}
Diederik~P. Kingma and Jimmy Ba.
\newblock Adam: a method for stochastic optimization.
\newblock {\em arXiv preprint arXiv:1412.6980}, 2017.

\bibitem{gans_comparison}
Mario Lucic et~al.
\newblock Are gans created equal? a large-scale study.
\newblock In S. Bengio, H. Wallach, H. Larochelle, K. Grauman, N. Cesa-Bianchi,
  and R. Garnett, editors, {\em Advances in Neural Information Processing
  Systems}, volume~31, pages 700--709. Curran Associates, Inc., 2018.

\bibitem{mathiasen2020fast}
Alexander Mathiasen and Frederik Hvilshøj.
\newblock Fast fr\'echet inception distance.
\newblock {\em arXiv preprint arXiv:2009.14075}, 2020.

\bibitem{mathieu2016deep}
Michael Mathieu, Camille Couprie, and Yann LeCun.
\newblock Deep multi-scale video prediction beyond mean square error.
\newblock {\em arXiv preprint arXiv:1511.05440}, 2016.

\bibitem{Menon_2020_CVPR}
Sachit Menon et~al.
\newblock Pulse: Self-supervised photo upsampling via latent space exploration
  of generative models.
\newblock In {\em Proceedings of the IEEE Conference on Computer Vision and
  Pattern Recognition}, June 2020.

\bibitem{mirza2014conditional}
Mehdi Mirza and Simon Osindero.
\newblock Conditional generative adversarial nets.
\newblock {\em arXiv preprint arXiv:1411.1784}, 2014.

\bibitem{ronneberger2015unet}
O. Ronneberger, P.Fischer, and T. Brox.
\newblock U-net: convolutional networks for biomedical image segmentation.
\newblock In {\em Medical Image Computing and Computer-Assisted Intervention},
  volume 9351 of {\em LNCS}, pages 234--241. Springer, 2015.

\bibitem{Seitzer2020FID}
Maximilian Seitzer.
\newblock pytorch-fid: fid score for pytorch.
\newblock \url{https://github.com/mseitzer/pytorch-fid}, August 2020.
\newblock Version 0.1.1.

\bibitem{song2021scorebased}
Yang Song et~al.
\newblock Score-based generative modeling through stochastic differential
  equations.
\newblock In {\em International Conference on Learning Representations}, 2021.

\bibitem{denoising_comaprison}
Rini Thakur, R.N. Yadav, and Lalita Gupta.
\newblock State-of-art analysis of image denoising methods using convolutional
  neural networks.
\newblock {\em IET Image Processing}, 13, 08 2019.

\bibitem{TIAN2020251}
Chunwei Tian et~al.
\newblock Deep learning on image denoising: an overview.
\newblock {\em Neural Networks}, 131:251 -- 275, 2020.

\bibitem{Tonolini2020VariationalIF}
Francesco Tonolini et~al.
\newblock Variational inference for computational imaging inverse problems.
\newblock {\em arXiv preprint arXiv:1904.06264}, 2020.

\bibitem{Vaksman_2020_CVPR_Workshops}
Gregory Vaksman, Michael Elad, and Peyman Milanfar.
\newblock Lidia: lightweight learned image denoising with instance adaptation.
\newblock In {\em The IEEE Conference on Computer Vision and Pattern
  Recognition Workshops}, June 2020.

\bibitem{whang2020approximate}
Jay Whang, Erik~M. Lindgren, and Alexandros~G. Dimakis.
\newblock Approximate probabilistic inference with composed flows.
\newblock {\em arXiv preprint arXiv:2002.11743}, 2020.

\bibitem{dsganICLR2019}
Dingdong Yang et~al.
\newblock Diversity-sensitive conditional generative adversarial networks.
\newblock In {\em Proceedings of the International Conference on Learning
  Representations}, 2019.

\bibitem{LSUNdataset}
Fisher Yu et~al.
\newblock Lsun: construction of a large-scale image dataset using deep learning
  with humans in the loop.
\newblock {\em arXiv preprint arXiv:1506.03365}, 2015.

\bibitem{yu2019deep}
Songhyun Yu, Bumjun Park, and Jechang Jeong.
\newblock Deep iterative down-up cnn for image denoising.
\newblock In {\em Proceedings of the IEEE Conference on Computer Vision and
  Pattern Recognition Workshops}, 2019.

\bibitem{Zhang_DnCnn_2017}
Kai Zhang et~al.
\newblock Beyond a gaussian denoiser: residual learning of deep cnn for image
  denoising.
\newblock {\em IEEE Transactions on Image Processing}, 26(7):3142--3155, 2017.

\bibitem{Zhang_FFDNet_2018}
Kai Zhang, Wangmeng Zuo, and Lei Zhang.
\newblock Ffdnet: toward a fast and flexible solution for cnn-based image
  denoising.
\newblock {\em IEEE Transactions on Image Processing}, 27(9):4608--4622, 2018.

\end{thebibliography}
}
\clearpage
\begin{appendices}
\section{Generator (Denoiser) Architecture and Full Framework Schematic}\label{generator_architecture}
\begin{figure*}[t]
    \centering
    \includegraphics[width=0.8\textwidth]{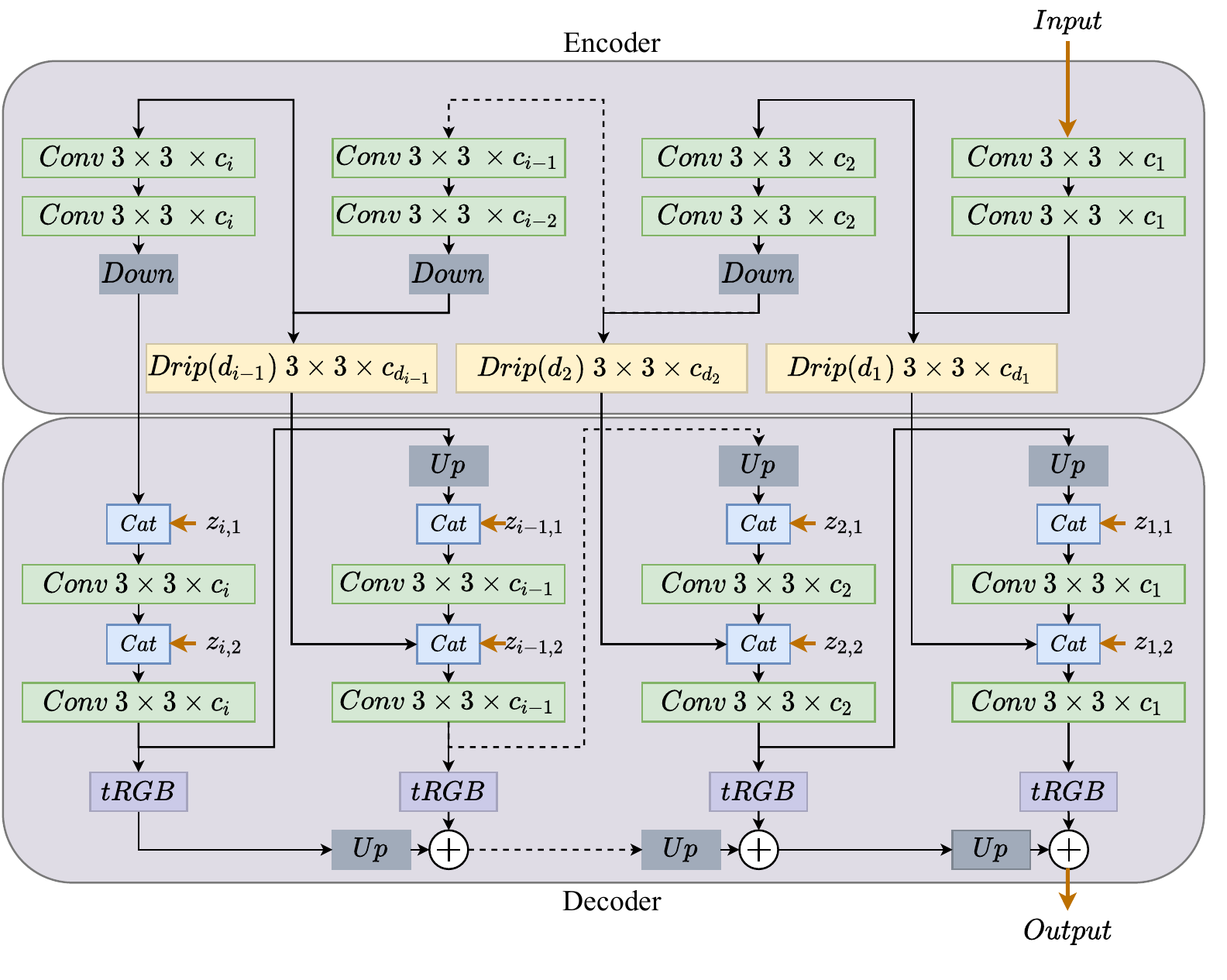}
    \caption{Our proposed generator architecture.
    An input noisy image is passed through an encoder of $i$ doubly-blocked convolutional layers and downsampled after each (except for the first block).
The downsampling operation is performed by a stride of $2$ in the preceding convolution layer.
The result of each doubly-blocked layer is then passed through a Drip, which is a feed-forward CNN (in the figure, $d_{k}$ and $c_{d_{k}}$ denote the number of layers and the number of output channels of each layer in drip $k$, respectively).
Each of these drips extracts features that are limited to a certain receptive field, which are then passed to the neighboring decoder block through concatenation.
This further assists in reconstructing the RGB result of the corresponding scale, especially at higher scales.
The decoder builds the reconstructed image scale by scale, using features aggregated from previous layers of the decoder and from the drip injections. 
Noise injections are performed in the decoder's pipeline, where a noisy ``image'', denoted as $z_{k,1}$ and $z_{k,2}$ for each $1\leq k\leq i$, is concatenated as another feature map of the next layer's input.
All convolutional layers, except for $tRGB$, are coupled with Leaky ReLU activation functions with a slope of $\alpha=0.2$ for negative values.
$tRGB$ is a simple convolution operation with output channels being equal to the number of channels of the input image (3 for RGB images).
All up-sampling operations are performed with nearest-neighbor interpolation.}
    \label{fig:generator}
\end{figure*}
\begin{figure*}[t]
    \centering
    \includegraphics{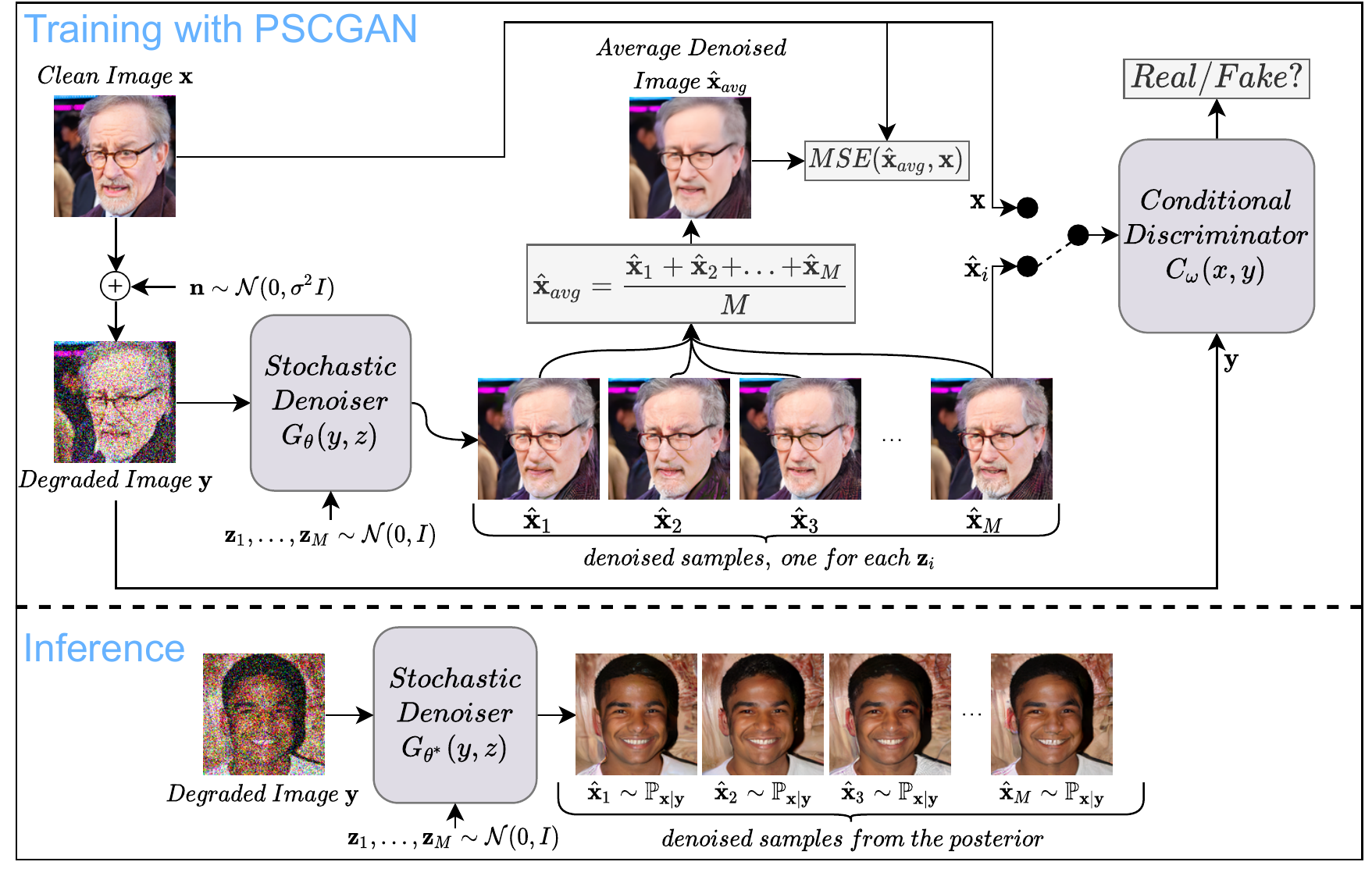}
    \caption{A schematic diagram describing our proposed method at both training time and inference time. During training, the denoiser receives a noisy image $\boldy$ and outputs many possible denoised candidates $\hat{\boldx}_{1}, \hat{\boldx}_{2},\hdots,\hat{\boldx}_{M}$. One of these candidates is being fed to a conditional discriminator, so as to drive the denoiser to produce images with high perceptual quality. At the same time, all of the denoised candidates are being averaged, and the result, $\hat{\boldx}_{avg}$, is forced to be close to the clean image $\boldx$ (in MSE). During inference, the denoiser receives a noisy input and produces many \emph{different} denoised candidates with high perceptual quality, as many as the number of provided latent noises $\boldz_i$.}
    \label{fig:full_schematic}
\end{figure*}
Inspired by StyleGAN2~\cite{Karras_2020_CVPR} and UNet~\cite{ronneberger2015unet}, our conditional generator (denoiser) is an encoder-decoder deep neural network, as shown in \autoref{fig:generator}.
The decoder builds the output image scale by scale by adding residual information at each stage to the up-sampled RGB output of the previous one.
This approach was proposed in StyleGAN2 to sidestep the shortcomings of progressive growing~\cite{karras2018progressive,Karras_2020_CVPR}, a training methodology in which the number of layers in the generator and the discriminator increases during training.
In the newly proposed scheme, both are trained end-to-end with all resolutions included, which significantly eases the training procedure but still enforces the decoder to progressively synthesize the output image stage-by-stage by adding resolution-specific details at each level.
The encoder is analogous to a drip irrigation system.
It consists of a main pipeline that has several exits to independent convolutional neural networks (CNNs), denoted as \emph{Drips}, each of which reinforces its neighboring decoder block with low receptive field information.
The main pipeline is a deep CNN with high receptive field that progressively encodes the input image.
This method alleviates the task of the decoder, especially at higher scales where original pixel locations are crucial for distortion performance.

Since our denoiser is stochastic and meant to sample from the posterior distribution, it can be considered as a mapping from the latent distribution of the random noise to the posterior, as in all GAN based sampling solutions.
Instead of injecting a single noisy tensor to the first layer of our model, we inject noise at each scale of the decoder.
These convolutional layers operate as follows: for a given layer with $c$ input channels $\{x_{i}\}_{i=1}^{c}$ and a random noise input $z$ of the same size, the resulting output of the layer (before the activation function is applied) is 
\begin{equation}
    \sum_{i=1}^{c}h_{i}*x_{i}+h_{c+1}*z,
\end{equation}
where $\{h_{i}\}_{i=1}^{c+1}$ are the convolutional kernels of the layer (considering only one block of kernels that leads to one output feature map).
If one further assumes that $h_{c+1}=\alpha$ (a 1-by-1 kernel) for some scaling factor $\alpha$, this boils down to the noise injection scheme of StyleGAN~\cite{Karras_2019_CVPR} (each resulting feature map corresponds to a different scaling factor).
In addition, if one forces the scaling factors of all feature maps to be equal, this becomes the noise injection scheme of StyleGAN2~\cite{Karras_2020_CVPR}.
Thus, our scheme enlarges the hypothesis set of the convolution operating on $z$. 
We incorporate this idea by concatenating each noise injection to the next convolutional layer's input.
Consult \autoref{fig:generator} for clarifications on the denoiser's architecture, and \autoref{fig:full_schematic} for a full framework schematic diagram.
\section{Latent Adversarial Generator Implementation Details}\label{lag_implementation}
Expression (\textcolor{red}{6}) is the originally proposed optimization task of LAG~\cite{berthelot2020creating}.
Yet, the SISR results presented in LAG's paper are achieved with a tweaked version,
\begin{gather}\label{obj:lag_opt_revision}
\min_{\theta}\supf[]\expect{\boldx}{f(\boldx,0)}-\expect{\gtheta,\boldy}{f(\gtheta,R(\gtheta,\boldy))}\\
+\lambda\mathbb{E}_{\boldx,\boldy}\left[\norm{P(\boldx,0)-P(G_{\theta}(0,\boldy),R(G_{\theta}(0,\boldy),\boldy))}_{2}^{2}\right],\nonumber
\end{gather}
in which, instead of $\boldy$, the critic receives $R(\gtheta,\boldy)$ as a second input, the pixel-wise absolute difference between the degraded image $\boldy$ and the corresponding degraded version of the generated image $\gtheta|\boldy$.
Such a tweak could also be adopted in the case of image denoising, for instance by defining $R(\gtheta,\boldy)$ to be the absolute difference between $\boldx$, the clean source of $\boldy$, and the corresponding denoised image $\gtheta|\boldy$.
However, such a revision deviates the posterior sampling goal, since the Wasserstein-1 distance~\cite{wgan} would consequently measure the deviation between the distributions $\mathbb{P}_{\boldx}$ and $\mathbb{P}_{\boldx|R(\gtheta,\boldy)}$.
We leave this, possibly beneficial, approach for future research.

In PSCGAN the critic receives $\boldy$ as its second input, and thus, for a fair comparison, we implement LAG in the same fashion instead of applying the aforementioned tweak.
Our choice also aligns with optimization task (\textcolor{red}{6}) (since the critic receives $\boldy$) and therefore leads to a more direct evaluation of it.

It is important to note that expression (\textcolor{red}{6}) is highly dependent on the choice of $P(\cdot,\cdot)$.
While many choices are possible, we find that in our case choosing $P(x,y)=x$ leads to superior results, both in the FID and the PSNR performance measures.
$P(x,y)=x$ means that we measure the distortion between $\boldx$ and $G_{\theta}(0,\boldy)$ in expression (\textcolor{red}{6}) with the MSE loss, operating in the RGB pixel space of the image instead of operating in an intermediate feature space.
While the sampled images attained at $\sigma_{\boldz}=1$ should achieve higher perceptual quality (regardless of the choice of $P(\cdot,\cdot)$), this choice, quite conveniently, also allows for a fair PSNR comparison between PSCGAN and LAG, since the images produced by LAG at $\boldz=0$ are now directly aimed to optimize the MSE.
We refer to this version of LAG as \emph{Ours-LAG} in the experimental evaluations, so as to emphasize that our choices deviate quite significantly from the original implementation of LAG (a different inverse problem, different architectures, and different loss).
\section{Training}\label{training}
\subsection{Data Splits}
\begin{itemize}[leftmargin=*]
    \item FFHQ~\cite{Karras_2019_CVPR} thumbnails contains 70,000 images.
    We use images 3000-4999 for testing and the rest for training.
    \item LSUN Bedroom~\cite{LSUNdataset} contains 3,037,042 images.
    We randomly pick 100,000 and 4,000 non-overlapping images for training and testing, respectively.
    \item LSUN Church~\cite{LSUNdataset} contains 126,227 images.
    We randomly pick 100,000 and 4,000 non-overlapping images for training and testing, respectively.
\end{itemize}
\subsection{Preprocessing}
Our model assumes an input image of size $128\times 128$, and since the images in both LSUN data sets are of larger size in both axes, we first center crop each image while keeping the smaller dimension fixed, and then resize the resulting square image to the desired size through interpolation.
All images in the FFHQ thumbnails data set are already of size $128\times 128$, and therefore do not require augmentation.
Finally, we use random horizontal flip during training in all data sets.
\subsection{Hyperparameters}
PSCGAN (and consequently PSCGAN-A) is trained with the default hyperparameters given in \textcolor{red}{Algorithm 1}.
Note that we evaluate the penalty term on the first $PB=8$ samples of each mini-batch of $B=32$ samples, and approximate $\mathbb{E}\left[G_{\theta}(\boldz,\boldy)|\boldy\right]$ by averaging $M=8$ generated samples for a given noisy image $\boldy$.
While this choice of $M$ may seem too small to evaluate the expectation of the posterior, it is sufficient to demonstrate the novelty of PSCGAN.

All other algorithms are also trained with a batch size of $32$ and the Adam optimizer~\cite{kingma2017adam}.
LAG is trained with a learning rate of $2.5\cdot 10^{-4}$, and the Adam hyperparameters are $\beta_{1}=0,\beta_{2}=0.99$ (similar to PSCGAN).
DnCNN and Ours-MSE are trained to optimize the MSE loss, the former with a learning rate of $10^{-3}$ and the latter with a learning rate of $5\cdot 10^{-4}$.
The Adam hyperparameters for both methods are $\beta_{1}=0.9,\beta_{2}=0.99$.

The full implementation of all methods and the checkpoints that reproduce the results reported in this paper are publicly available. 
Our implementations are based on PyTorch and PyTorch Lightning~\cite{falcon2019pytorch}.
\section{LSUN Data Sets' Visual Results}\label{appendix:lsun_visual_results}
\begingroup
\setlength{\tabcolsep}{0pt} 
\renewcommand{\arraystretch}{0} 
    \begin{figure*}[t]
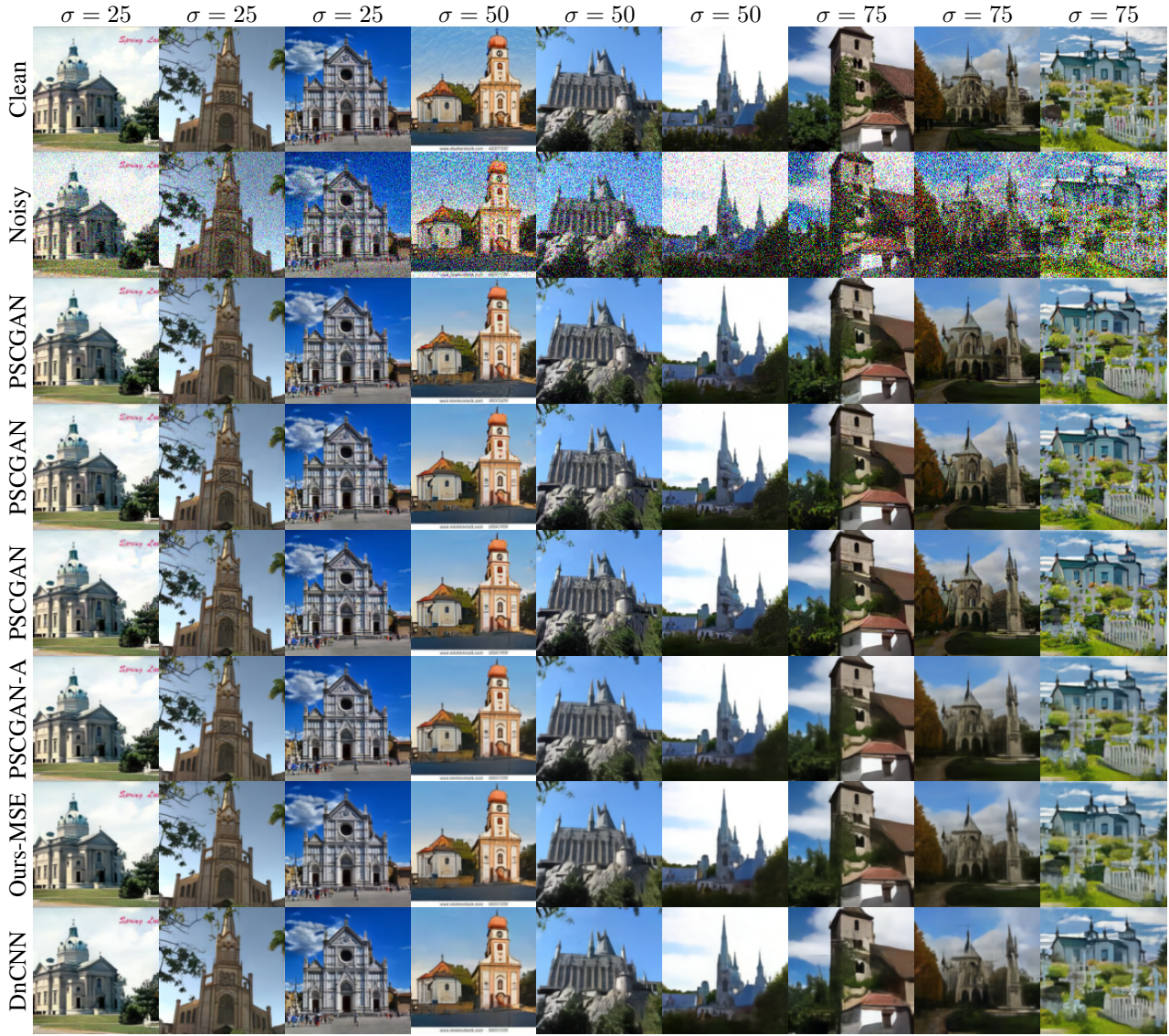

        \centering
        \begin{tabular}{p{0.02\textwidth} c c c c c c c c c}
        &$\sigma=25$&$\sigma=25$&$\sigma=25$&$\sigma=50$&$\sigma=50$&$\sigma=50$&$\sigma=75$&$\sigma=75$&$\sigma=75$\\
        \rule{0pt}{0.4ex}&&&&&&&&&\\
        \centered{\begin{turn}{90}Clean\end{turn}}&
        \real{25}{church}{0}{0}&
        \real{25}{church}{0}{2}&
        \real{25}{church}{0}{4}&
        \real{50}{church}{3}{4}&
        \real{50}{church}{4}{5}&
        \real{50}{church}{8}{2}&
        \real{75}{church}{1}{3} &
        \real{75}{church}{1}{5}&
        \real{75}{church}{1}{6}\\
        \centered{\begin{turn}{90}Noisy\end{turn}}&
        \noisy{25}{church}{0}{0}&
        \noisy{25}{church}{0}{2}&
        \noisy{25}{church}{0}{4}&
        \noisy{50}{church}{3}{4}&
        \noisy{50}{church}{4}{5}&
        \noisy{50}{church}{8}{2}&
        \noisy{75}{church}{1}{3}&
        \noisy{75}{church}{1}{5}&
        \noisy{75}{church}{1}{6}\\
        \centered{\begin{turn}{90}PSCGAN\end{turn}}&
        \oursslikely{25}{church}{0}{0}{00}&
        \oursslikely{25}{church}{0}{2}{00}&
        \oursslikely{25}{church}{0}{4}{00}&
        \oursslikely{50}{church}{3}{4}{00}&
        \oursslikely{50}{church}{4}{5}{00}&
        \oursslikely{50}{church}{8}{2}{00}&
        \oursslikely{75}{church}{1}{3}{00}&
        \oursslikely{75}{church}{1}{5}{00}&
        \oursslikely{75}{church}{1}{6}{00}\\
        \centered{\begin{turn}{90}PSCGAN\end{turn}}&
        \oursslikely{25}{church}{0}{0}{01}&
        \oursslikely{25}{church}{0}{2}{01}&
        \oursslikely{25}{church}{0}{4}{01}&
        \oursslikely{50}{church}{3}{4}{01}&
        \oursslikely{50}{church}{4}{5}{01}&
        \oursslikely{50}{church}{8}{2}{01}&
        \oursslikely{75}{church}{1}{3}{01}&
        \oursslikely{75}{church}{1}{5}{01}&
        \oursslikely{75}{church}{1}{6}{01}\\
        \centered{\begin{turn}{90}PSCGAN\end{turn}}&
        \oursslikely{25}{church}{0}{0}{01}&
        \oursslikely{25}{church}{0}{2}{01}&
        \oursslikely{25}{church}{0}{4}{01}&
        \oursslikely{50}{church}{3}{4}{01}&
        \oursslikely{50}{church}{4}{5}{01}&
        \oursslikely{50}{church}{8}{2}{01}&
        \oursslikely{75}{church}{1}{3}{01}&
        \oursslikely{75}{church}{1}{5}{01}&
        \oursslikely{75}{church}{1}{6}{01}\\
        \centered{\begin{turn}{90}PSCGAN-A\end{turn}}&
        \oursa{25}{church}{0}{0} &
        \oursa{25}{church}{0}{2} &
        \oursa{25}{church}{0}{4} &
        \oursa{50}{church}{3}{4}&
        \oursa{50}{church}{4}{5}&
        \oursa{50}{church}{8}{2}&
        \oursa{75}{church}{1}{3}&
        \oursa{75}{church}{1}{5}&
        \oursa{75}{church}{1}{6}\\
        \centered{\begin{turn}{90}Ours-MSE\end{turn}}&
        \oursmse{25-church}{0-0}&
        \oursmse{25-church}{0-2}&
        \oursmse{25-church}{0-4}&
        \oursmse{50-church}{3-4}&
        \oursmse{50-church}{4-5}&
        \oursmse{50-church}{8-2}&
        \oursmse{75-church}{1-3}&
        \oursmse{75-church}{1-5}&
        \oursmse{75-church}{1-6}\\
        \centered{\begin{turn}{90}DnCNN\end{turn}}&
        \dncnn{25-church}{0-0}&
        \dncnn{25-church}{0-2}&
        \dncnn{25-church}{0-4}&
        \dncnn{50-church}{3-4}&
        \dncnn{50-church}{4-5}&
        \dncnn{50-church}{8-2}&
        \dncnn{75-church}{1-3}&
        \dncnn{75-church}{1-5}&
        \dncnn{75-church}{1-6}
        \end{tabular}
        \caption{Denoising results on the LSUN Church outdoor test set produced by several methods. For each image we show three different outcomes of PSCGAN, each attained by injecting noise with standard deviation of $\sigma_{\boldz}=0.75$.
        In this case, PSCGAN-A averages 64 instances of PSCGAN, where each instance is attained with $\sigma_{\boldz}=1$ at inference time. Each model is trained on the LSUN Church outdoor training set to denoise a specific noise level ($25, 50$ or $75$).}
        \label{fig:collage_church}
    \end{figure*}
\endgroup

\begingroup
\setlength{\tabcolsep}{0pt} 
\renewcommand{\arraystretch}{0} 
    \begin{figure*}[t]
        \centering
        \begin{tabular}{p{0.02\textwidth} c c c c c c c c c}
        &$\sigma=25$&$\sigma=25$&$\sigma=25$&$\sigma=50$&$\sigma=50$&$\sigma=50$&$\sigma=75$&$\sigma=75$&$\sigma=75$\\
        \rule{0pt}{0.4ex}&&&&&&&&&\\
        \centered{\begin{turn}{90}Clean\end{turn}}&
        \real{25}{bed}{6}{0}&
        \real{25}{bed}{6}{1}&
        \real{25}{bed}{6}{2}&
        \real{50}{bed}{9}{0}&
        \real{50}{bed}{9}{1}&
        \real{50}{bed}{9}{3}&
        \real{75}{bed}{0}{2}&
        \real{75}{bed}{0}{4}&
        \real{75}{bed}{1}{0}\\
        \centered{\begin{turn}{90}Noisy\end{turn}}&
        \noisy{25}{bed}{6}{0}&
        \noisy{25}{bed}{6}{1}&
        \noisy{25}{bed}{6}{2}&
        \noisy{50}{bed}{9}{0}&
        \noisy{50}{bed}{9}{1}&
        \noisy{50}{bed}{9}{3}&
        \noisy{75}{bed}{0}{2}&
        \noisy{75}{bed}{0}{4}&
        \noisy{75}{bed}{1}{0}\\
        \centered{\begin{turn}{90}PSCGAN\end{turn}}&
        \oursslikely{25}{bed}{6}{0}{00}&
        \oursslikely{25}{bed}{6}{1}{00}&
        \oursslikely{25}{bed}{6}{2}{00}&
        \oursslikely{50}{bed}{9}{0}{00}&
        \oursslikely{50}{bed}{9}{1}{00}&
        \oursslikely{50}{bed}{9}{3}{00}&
        \oursslikely{75}{bed}{0}{2}{00}&
        \oursslikely{75}{bed}{0}{4}{00}&
        \oursslikely{75}{bed}{1}{0}{00} \\
        \centered{\begin{turn}{90}PSCGAN\end{turn}}&
        \oursslikely{25}{bed}{6}{0}{01}&
        \oursslikely{25}{bed}{6}{1}{01}&
        \oursslikely{25}{bed}{6}{2}{01}&
        \oursslikely{50}{bed}{9}{0}{01}&
        \oursslikely{50}{bed}{9}{1}{01}&
        \oursslikely{50}{bed}{9}{3}{01}&
        \oursslikely{75}{bed}{0}{2}{01}&
        \oursslikely{75}{bed}{0}{4}{01}&
        \oursslikely{75}{bed}{1}{0}{01} \\
        \centered{\begin{turn}{90}PSCGAN\end{turn}}&
        \oursslikely{25}{bed}{6}{0}{02}&
        \oursslikely{25}{bed}{6}{1}{02}&
        \oursslikely{25}{bed}{6}{2}{02}&
        \oursslikely{50}{bed}{9}{0}{02}&
        \oursslikely{50}{bed}{9}{1}{02}&
        \oursslikely{50}{bed}{9}{3}{02}&
        \oursslikely{75}{bed}{0}{2}{02}&
        \oursslikely{75}{bed}{0}{4}{02}&
        \oursslikely{75}{bed}{1}{0}{02}\\
        \centered{\begin{turn}{90}PSCGAN-A\end{turn}}&
        \oursa{25}{bed}{6}{0}&
        \oursa{25}{bed}{6}{1}&
        \oursa{25}{bed}{6}{2}&
        \oursa{50}{bed}{9}{0}&
        \oursa{50}{bed}{9}{1}&
        \oursa{50}{bed}{9}{3}&
        \oursa{75}{bed}{0}{2}&
        \oursa{75}{bed}{0}{4}&
        \oursa{75}{bed}{1}{0}\\
        \centered{\begin{turn}{90}Ours-MSE\end{turn}}&
        \oursmse{25-bed}{6-0}&
        \oursmse{25-bed}{6-1}&
        \oursmse{25-bed}{6-2}&
        \oursmse{50-bed}{9-0}&
        \oursmse{50-bed}{9-1}&
        \oursmse{50-bed}{9-3}&
        \oursmse{75-bed}{0-2}&
        \oursmse{75-bed}{0-4}&
        \oursmse{75-bed}{1-0}\\
        \centered{\begin{turn}{90}DnCNN\end{turn}}&
        \dncnn{25-bed}{6-0}&
        \dncnn{25-bed}{6-1}&
        \dncnn{25-bed}{6-2}&
        \dncnn{50-bed}{9-0}&
        \dncnn{50-bed}{9-1}&
        \dncnn{50-bed}{9-3}&
        \dncnn{75-bed}{0-2}&
        \dncnn{75-bed}{0-4}&
        \dncnn{75-bed}{1-0}
        \end{tabular}
        \caption{Denoising results on the LSUN Bedroom test set produced by several methods. For each image we show three different outcomes of PSCGAN, each attained by injecting noise with standard deviation of $\sigma_{\boldz}=0.75$.
        In this case, PSCGAN-A averages 64 instances of PSCGAN, where each instance is attained with $\sigma_{\boldz}=1$ at inference time. Each model is trained on the LSUN Bedroom training set to denoise a specific noise level ($25, 50$ or $75$).}
        \label{fig:collage_bed}
    \end{figure*}
\endgroup

In \autoref{fig:collage_church} and \autoref{fig:collage_bed} we illustrate the visual quality of several denoised images produced by our method and by other MSE based methods on the LSUN Church outdoor and the LSUN Bedroom test sets.
As can be seen, our model produces denoised images with high perceptual quality, although in these data sets it is harder to notice the perceptual quality difference with the naked eye (since the images are compressed).
\end{appendices}
\end{document}